\documentclass{article}

\usepackage{overpic}
\usepackage[final, nonatbib]{neurips_2023}
\usepackage{subcaption}
\usepackage{tabularx}
\usepackage{graphicx}

\usepackage[utf8]{inputenc} 
\usepackage[T1]{fontenc}    
\usepackage{hyperref}       
\usepackage{url}            
\usepackage{booktabs}       
\usepackage{amsfonts}       
\usepackage{nicefrac}       
\usepackage{microtype}      
\usepackage{xcolor}         
\usepackage{algorithm}
\usepackage[noend]{algpseudocode}
\usepackage{ragged2e}
\usepackage{graphicx}
\usepackage{amsmath}
\usepackage{diagbox}
\usepackage{multirow}
\usepackage[style=numeric, maxbibnames=99]{biblatex}       

\usepackage{amssymb}
\usepackage{amsmath,amsfonts}
\usepackage{amsopn}
\usepackage{bm} 
\usepackage{multirow}
\usepackage{flushend}
\usepackage{tabularx}

\newcommand{\vct}[1]{\boldsymbol{#1}} 
\newcommand{\mat}[1]{\boldsymbol{#1}} 

\newcommand{\field}[1]{\mathbb{#1}}
\newcommand{\R}{\field{R}} 

\newcommand{\ProbOpr}[1]{\mathbb{#1}}

\newcommand{\expect}[2]{%
\ifthenelse{\equal{#2}{}}{\ProbOpr{E}_{#1}}
{\ifthenelse{\equal{#1}{}}{\ProbOpr{E}\left[#2\right]}{\ProbOpr{E}_{#1}\left[#2\right]}}} 

\newcommand{\vy}{\vct{y}}

\newcommand{\mS}{\mat{S}}

\newcommand{\mM}{\mat{M}}

\newcommand{\mP}{\mat{P}}

\newcommand{\mX}{\mat{X}}

\newcommand{\eat}[1]{}

\title{Segment Anything Model (SAM) Enhances Pseudo- Labels for Weakly Supervised Semantic Segmentation}

\author{
  Tianle Chen $^{1}$\thanks{Equal contributions.} \quad 
Zheda Mai $^{1}$\footnotemark[1]\quad 
  Ruiwen Li$^{2}$\quad
  Wei-lun Chao $^{1}$ \\
  $^{1}$The Ohio State University\\
  $^{2}$EAIGLE Inc\\
  \texttt{(mai.145, chen.9471)@osu.edu, ruiwen@eaigle.com, chao.209@osu.edu}
}

\begin{document}

\maketitle

\begin{abstract}
Weakly supervised semantic segmentation (WSSS) aims to bypass the need for laborious pixel-level annotation by using only image-level annotation. Most existing methods rely on Class Activation Maps (CAM) to derive pixel-level pseudo-labels and use them to train a fully supervised semantic segmentation model. Although these pseudo-labels are class-aware, indicating the coarse regions for particular classes, they are not object-aware and fail to delineate accurate object boundaries. To address this, we introduce a simple yet effective method harnessing the Segment Anything Model (SAM), a class-agnostic foundation model capable of producing fine-grained instance masks of objects, parts, and subparts. We use CAM pseudo-labels as cues to select and combine SAM masks, resulting in high-quality pseudo-labels that are both class-aware and object-aware. Our approach is highly versatile and can be easily integrated into existing WSSS methods without any modification. Despite its simplicity, our approach shows consistent gain over the state-of-the-art WSSS methods on both PASCAL VOC and MS-COCO datasets. 

\end{abstract}

\section{Introduction}%
Semantic segmentation, a task aiming to assign a semantic label to each image pixel~\cite{SS_survey}, has found wide applications in various fields, such as medical imaging~\cite{asgari2021deep}, remote sensing~\cite{yuan2021review} and autonomous driving~\cite{feng2020deep}. The success of deep learning techniques and the availability of large-scale pixel-level annotations have greatly boosted the performance of semantic segmentation in recent years~\cite{lateef2019survey}. However, acquiring pixel-level annotations is daunting due to its laborious and costly nature. As an alternative, weakly supervised semantic segmentation (WSSS) seeks to train a segmentation model with cheaper yet weaker annotations such as bounding boxes~\cite{oh2021backgroundBox2,wsss_box1,wsss_box2,wsss_box3}, scribbles~\cite{lin2016scribblesup}, points~\cite{wsss_point1,wsss_point2}, and image-level class labels~\cite{wsss4,wsss3,wsss2,wsss1}. Among existing approaches, image-level WSSS has gained widespread popularity due to the abundance of image-level annotations online or in various vision datasets~\cite{russakovsky2015imagenet,COCO} and the availability of strong pre-trained classifiers~\cite{resnet, vgg}.



\begin{figure}[tb]
\centering

    \includegraphics[width=1\linewidth]{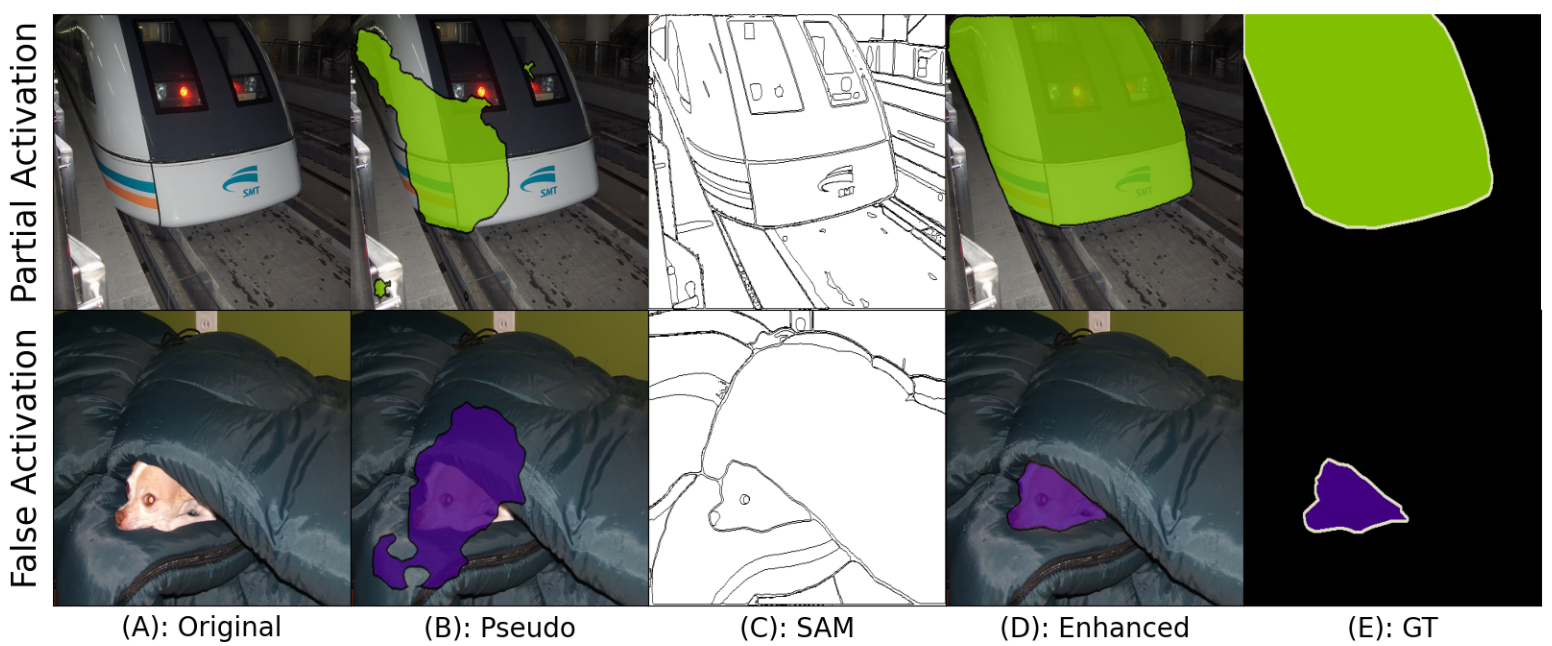}
    \vskip -2pt
    \caption{\small \textbf{Illustration of how SAM addresses partial and false activation} on PASCAL VOC 2012 train set: (A) original images; (B) pseudo-labels generated by a SOTA image-level WSSS method, CLIMS~\cite{xie2022clims}; (C) masks from SAM; (D) SAM enhanced pseudo-labels; (E) ground-truth labels. }
    \label{fig:demo}
    \vskip-14pt
\end{figure}

As image-level labels do not provide location information for each object class, most of the existing WSSS methods leverage Class Activation Maps (CAM)~\cite{cam} 
to derive location cues. 
These approaches typically follow a four-stage learning process. First, they train a classification model with image-level labels.
Then, based on the intermediate feature maps and their weights to a class, CAMs are generated as the coarse estimate of the class location. 
Subsequently, the initial CAMs are refined with post-processing techniques, such as pixel affinity-based methods~\cite{affinitynet,li2023transcam} or saliency guidance \cite{salency2,saliency1,yao2020saliency2}, to create pixel-level pseudo-labels. Finally, a semantic segmentation model~\cite{chen2014semantic_Deeplabv1,chen2017deeplabV2} is trained using the pseudo-labels as pixel-level supervision. The efficacy of WSSS greatly relies on the accuracy of pseudo-labels. However, it is widely recognized that the CAM-derived pseudo-labels often suffer from \emph{partial activation}~\cite{ahn2018learning, wang2020self}, activating the most discriminative region instead of the entire object area, and \emph{false activation}~\cite{xie2022clims, jo2022recurseed}, wrongly activating the background around the object. (\autoref{fig:demo} shows partial and false activation examples.) \emph{In other words, CAM-derived pseudo-labels lack the awareness of objects, resulting in poor contours that drastically deviate from object boundaries.} 

In this paper, we investigate a novel approach to addressing this issue, directly incorporating object boundary information into pseudo-label generation. 
Concretely, we leverage the advent of segmentation foundation models \cite{wang2023seggpt, kirillov2023segment, zou2023segment}, in particular, the Segment Anything Model (SAM) \cite{kirillov2023segment}, which is capable of producing fine-grained, class-agnostic masks of objects, parts, or subparts. 
We hypothesize that the quality of the resulting pseudo-labels can be greatly enhanced by appropriately integrating the coarse class location from CAM and the object boundary information from SAM.



To this end, we propose \textbf{SAM Enhanced pseudo-labels} (\textbf{SEPL}). \textbf{SEPL} uses CAM-derived pseudo-labels for a particular class as the seed signals to select the most relevant masks from SAM. The union of these masks, which encompass both class and object information, is then treated as the enhanced pseudo-labels for training semantic segmentation models. More specifically, \textbf{SEPL} consists of two stages: mask assignment and mask selection (see \autoref{fig:frwk}). During \textbf{mask assignment}, each SAM mask is assigned to the class (of the image-level annotation) whose CAM-derived pseudo-labels have the largest intersection with the mask.
During \textbf{mask selection}, SAM masks with substantial overlap with the CAM-derived pseudo-labels are chosen to address \emph{false activation}, given that background masks typically manifest minimal overlap. 
Meanwhile, we also select SAM masks that substantially encompass the CAM-derived pseudo-labels, targeting the challenge of \emph{partial activation}. Given the precise alignment of SAM masks to object boundaries, we find substantial enhancements in mitigating partial and false activations in the existing pseudo-labels, as depicted in~\autoref{fig:demo}.

SEPL is remarkably versatile as it can be seamlessly integrated into existing WSSS methods without modifying the original methods. Despite its simplicity, SEPL achieves a notable improvement in the mean Intersection over Union (mIoU) of pseudo-labels and ground-truth labels compared to eleven state-of-the-art WSSS methods, with an average gain of 5.33\% and 3.12\% on the train set of PASCAL VOC 2012 dataset~\cite{everingham2010pascal} and MS COCO 2014~\cite{COCO}, respectively. As far as we know, this is the first study to investigate the potential of SAM in the context of WSSS. We hope this work will pave the way for applying segmentation foundation models in diverse computer vision applications.

\textbf{Related work.} Due to the page limit, we leave it in~\autoref{ssup_related}.

\section{SAM Enhanced Pseudo Labels}

\subsection{Preliminary}

Following the standard setup, each training image $\mX \in \R^{H\times W \times C}$ is associated with only an image-level label vector $\vy=\left[y_1, y_2, \ldots, y_K\right]^\top \in\{0,1\}^K$ for $K$ classes, where $y_k=1$ indicates the presence of class $k$ in $\mX$ and 0 otherwise. Upon training a classifier $f$ with this dataset, WSSS methods feed an image to $f$ and obtain the Class Activation Maps (CAM) $\mathcal{M}=\left[\mM_1, \cdots \mM_K\right]$, where $\mM_k \in R^{H\times W}$ highlights the discriminative image regions utilized by $f$ to identify class $k$. Post-processing techniques, such as AffinityNet~\cite{ahn2018learning} and IRNet~\cite{ahn2019weakly}, further refine $\mM_k$ to produce pseudo-labels $\mP \in \{0, 1, \cdots, K \}^{H\times W}$, where pixel is mapped to either a class label in $\{1, \cdots, K \}$ or 0 for background regions. $\mP_k \in \{0, k\}^{H \times W}$ represents the pseudo-labels for class $k$. Subsequently, a fully supervised semantic segmentation network (e.g., \cite{chen2017deeplab,chen2017deeplabV2}) is trained with $\mP$.
SAM takes an image $\mX \in \R^{H\times W \times C}$ as input and returns a list of masks capturing either a subpart, a part or an entire object: $\mathcal{S}=\left[\mS_0, \mS_1, \cdots, \mS_L\right]$ where $\mS_l \in \{0, 1\}^{H \times W}$ and $L$ is the number of masks. It is noteworthy that different images may receive various mask quantities from SAM. 

\begin{figure}[t]
\centering
\includegraphics[width=1\linewidth]{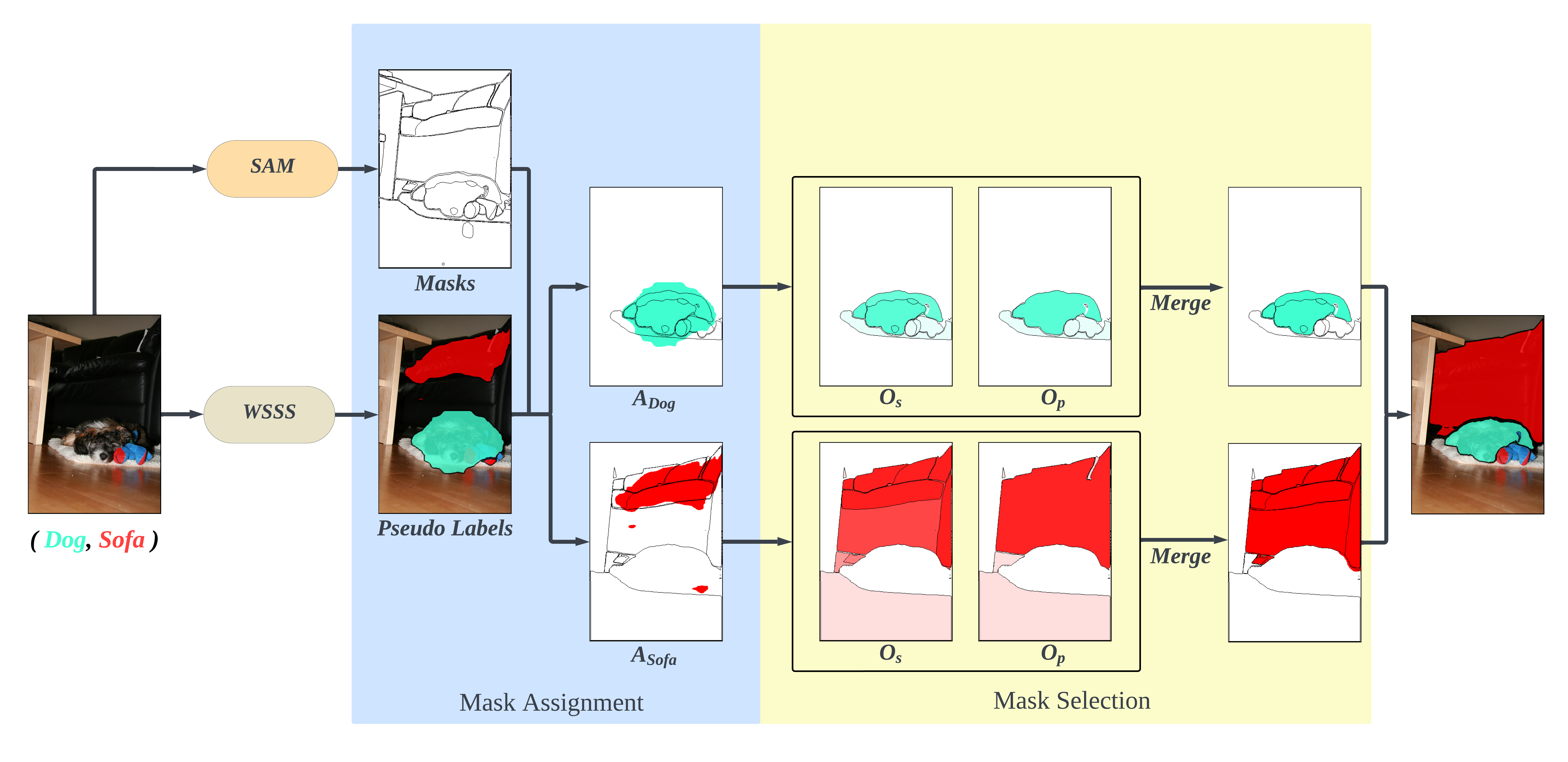}
    \vskip-5pt
    \caption{\small \textbf{Illustration of the SEPL pipeline.} SEPL comprises two stages, mask assignment and mask selection. Based on the intersection between each SAM mask and pseudo-labels, a mask is assigned to the class with the largest intersection. For each mask, two metrics are computed: $o_s$, the fraction of the mask overlapped by pseudo-labels, and $o_p$, the fraction of pseudo-labels overlapped by the mask. A mask is retained as an enhanced pseudo-label if either metric surpasses the designated threshold.}
        \label{fig:frwk}
    \vskip-15pt
\end{figure}
\vspace{-1mm}
\subsection{Approach}

While CAM-derived pseudo-labels are class-aware and identify discriminative regions for individual classes, they often fail to delineate accurate object boundaries. In contrast, SAM is able to precisely segment most parts or objects in a class-agnostic manner. Bridging their capabilities, we propose SAM Enhanced Pseudo Labels (SEPL) to harness the potential of SAM for pseudo-label enhancement. 

Our approach takes pseudo-labels $[\mP_1, \cdots, \mP_K]$ and SAM masks $[\mS_0, \cdots, \mS_L]$ as input and  returns a list of enhanced pseudo-labels $[\hat{\mP_1}, \cdots, \hat{\mP_K}]$. Specifically, it consists of two stages, mask assignment and mask selection as elucidated in~\autoref{fig:frwk}. During the mask assignment phase, we compute the intersection between each SAM mask $\mS_l$ and pseudo-labels $\mP_k$ for every class $k \in \{1, \cdots, K\}$. Each $\mS_l$ is assigned to the class with the largest intersection area and masks without any overlap with existing pseudo-labels are disregarded. After processing all SAM masks, we obtain a mask assignment list $A = [A_1, \cdots, A_K]$ where $A_k$ contains the masks assigned to class $k$. 

It is widely known that the CAM-derived pseudo-labels often suffer from false activation~\cite{xie2022clims, jo2022recurseed} and partial activation~\cite{ahn2018learning, wang2020self}. Our mask selection strategy aims to address them by selecting the most relevant masks based on the overlaps between SAM masks and pseudo-labels. False activation arises when pseudo-labels encapsulate the target object along with a marginal section of the surrounding background. Thus, the masks for the entire target object (or its parts) should predominantly align with the pseudo-labels, whereas masks for the background should exhibit minimal overlap with the pseudo-labels. To mitigate false activation, we select masks demonstrating extensive coverage by the pseudo-labels. Conversely, partial activation arises when pseudo-labels only cover the most discriminative part instead of the entire object. Therefore, if a mask covers the majority of the pseudo-labels, it is indicative that this mask likely represents the complete object and should be retained for enhanced pseudo-labels to address partial activation. 

Based on the intuitions mentioned above, we iterate through every mask assigned to class $k$. For each mask $\mS$, we compute $o_s$, the fraction of mask $\mS$ overlapped by pseudo-labels $\mP_k$, and $o_p$, the fraction of pseudo-labels $\mP_k$ overlapped by mask $\mS$. A mask $\mS$ is preserved as an enhanced pseudo-label if:

\begin{enumerate}
    \item $o_s > t_1$ where $t_1=0.5$ indicates at least 50\% of the mask is covered by the pseudo-labels
    \item $o_p > t_2$ where $t_2=0.85$ indicates at least 85\% of the pseudo-labels is covered by the mask
\end{enumerate}

If the initial pseudo-labels are not covered by any SAM masks, we will keep them unchanged in the enhanced pseudo-labels. The overall algorithm is summarized in Algorithm~\autoref{algo: sepl}. 
 



\begin{algorithm}
\small
\caption{SAM Enhanced Pseudo-Labels (SEPL) for One Image}
\hspace*{\algorithmicindent} \textbf{Input}: Pseudo labels $[\mP_1, \cdots, \mP_K]$, Masks $[\mS_0, \cdots, \mS_L]$, \\
\hspace*{\algorithmicindent} \hspace*{\algorithmicindent} \hspace{4mm} threshold $t_1=0.5$, threshold $t_2=0.85$ \\
\hspace*{\algorithmicindent} \textbf{Output} Enhanced pseudo-labels $[\hat{\mP_1}, \cdots, \hat{\mP_K}]$
\begin{algorithmic}
\\\hrulefill
\Procedure{SEPL}{$[\mP_1, \cdots, \mP_K]$, $[\mS_0, \cdots, \mS_L]$}
    \State $A = [A_1, \cdots, A_K]$ where $A_k = \{\}$ \algorithmiccomment{$A_k$ stores the masks assigned to class $k$} 
    \For{$l$ from $0$ to $L$}  \algorithmiccomment{Mask assignment} 
        \State $k^\star = \underset{k}{\arg\max} \; \text{Intersect}(\mS_l, \mP_k)$ 
        \State $A_{k^\star} \gets A_{k^\star} \cup \{\mS_l\}$
    \EndFor
    \For{$k$ from $1$ to $K$} \algorithmiccomment{Mask selection}
        \If{$\mP_k == \{0\}^{H \times W}$}
            \State \textbf{continue}
        \EndIf
        \State $tmp = \{\}$ \algorithmiccomment{$tmp$ stores enhanced pseudo-labels for class $k$}
        \For{each mask $S$ in $A_k$} 
            \State $o_s = \frac{\text{Intersect}(S, \mP_k)}{\text{nonzero\_area}(S)}$ \algorithmiccomment{fraction of mask $S$ covered by pseudo-labels  $\mP_k$}
            \State $o_p = \frac{\text{Intersect}(S, \mP_k)}{\text{nonzero\_area}(\mP_k)}$ \algorithmiccomment{fraction of pseudo-labels $\mP_k$ covered by mask $S$}
                \If{$o_s > t_1$ or $o_p > t_2$}
                    \State $tmp \gets tmp \cup \{S\}$
                \EndIf
        \EndFor
        \If{$tmp == \{\}$} 
            \State $tmp \gets tmp \cup \{\mP_k\}$
        \EndIf
        \State $\hat{\mP_k} \gets$ Merge masks in $tmp$ with element-wise OR and assign $k$ to nonzero elements

    \EndFor
    
\EndProcedure
\end{algorithmic}
\label{algo: sepl}
\end{algorithm}
\vspace{-2mm}

\section{Experiment}



\subsection {Experimental setup}
\paragraph{Datasets and Evaluation Metric}
We evaluate our proposed framework on the PASCAL VOC 2012~\cite{everingham2010pascal} and MS COCO 2014 dataset~\cite{lin2014microsoft}. The dataset details can be found in Appendix~\autoref{app: data}.  We only used image-level ground-truth labels during pseudo-labels generation. The mean Intersection over Union (mIoU) is adopted as the evaluation metric for all experiments. To demonstrate the quality of the pseudo-labels, we evaluate them on the VOC and COCO training set.

\paragraph{Pseudo Labels} In our experiments, we generate pseudo-labels using several SOTA WSSS methods, including: 
Recurseed~\cite{jo2022recurseed}, {L2G}~\cite{jiang2022l2g}, {CLIPES}~\cite{Lin_2023_CVPR_CLIPES}, {RCA}~\cite{zhou2022regional_rca}, {EPS}~\cite{lee2021railroad_eps}, {CLIMS}~\cite{xie2022clims}, {TransCAM}~\cite{li2023transcam}, {PPC}~\cite{wseg}, {SIPE}~\cite{SIPE}, and {PuzzleCAM}~\cite{jo2021puzzle}. The detailed introductions of them can be found in Appendix~\autoref{app: baseline}. 
\paragraph{Implementation Details} SAM masks are generated with the official code~\cite{kirillov2023segment}. The hyperparameters used for inference can be found in Appendix~\autoref{app: impl}. $t_1$ and $t_2$ by default are set to 0.5 and 0.85 respectively. We use Deeplab V2 (ResNet-101)~\cite{chen2017deeplab} as the fully supervised semantic segmentation model.
\subsection{Quantitative Evaluation and Comparison}
\begin{figure}[tb]
\centering
    \includegraphics[width=1\linewidth]{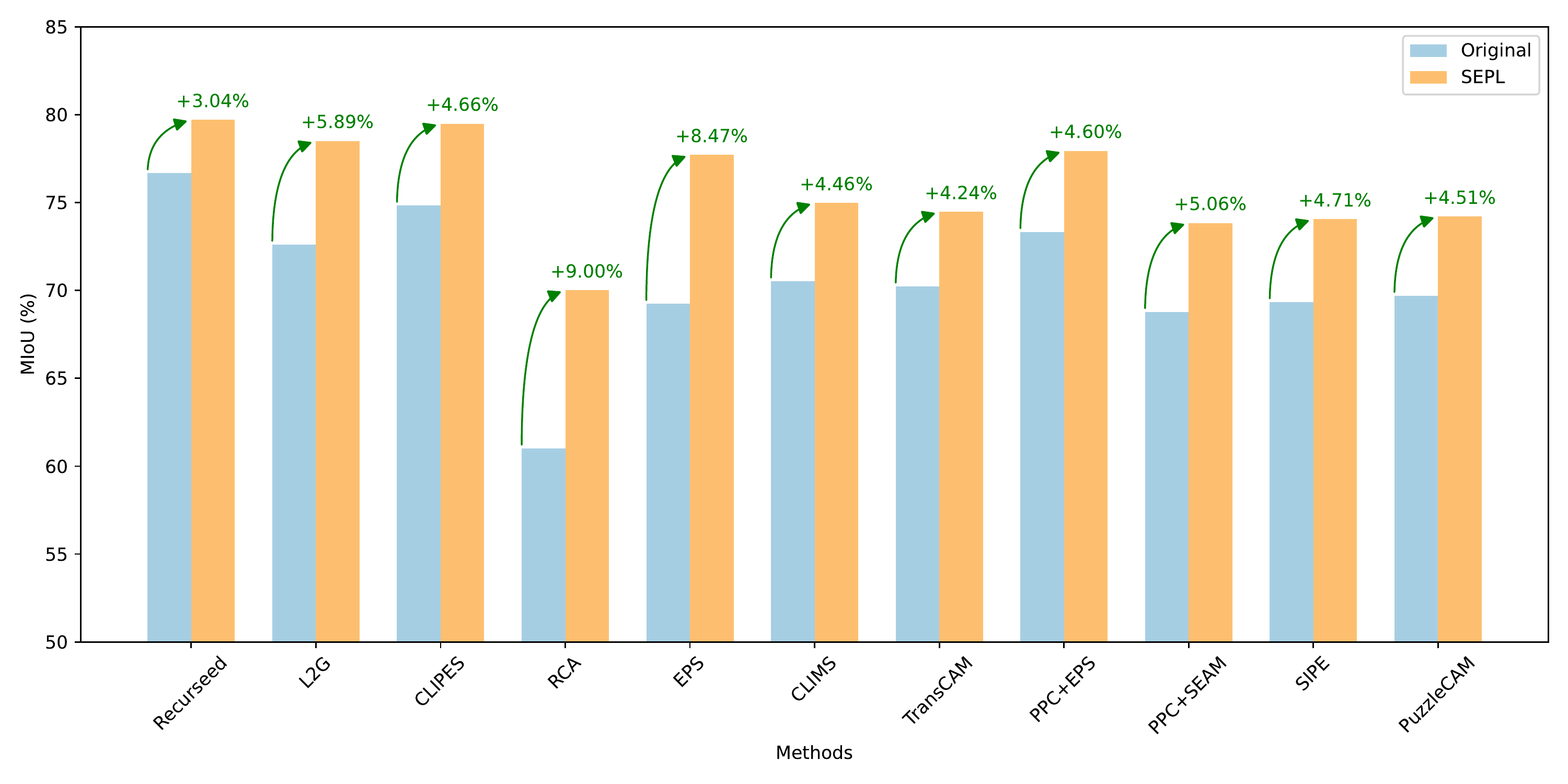}
    \vskip-8pt
    \caption{\small Pseudo labels quality on PASCAL VOC 2012. Original: pseudo-labels from original SOTA WSSS methods. SEPL: pseudo-labels enhanced by SEPL. The improvement after enhancement is indicated in green.}
    \label{fig:pseudo_label_voc}
    \vspace{-2mm}
\end{figure}

\begin{figure}[ht]
\begin{minipage}{.48\textwidth}
    \centering
    \vspace{0.5mm}
    \includegraphics[width=\linewidth, trim=0 0 0 1cm, clip]{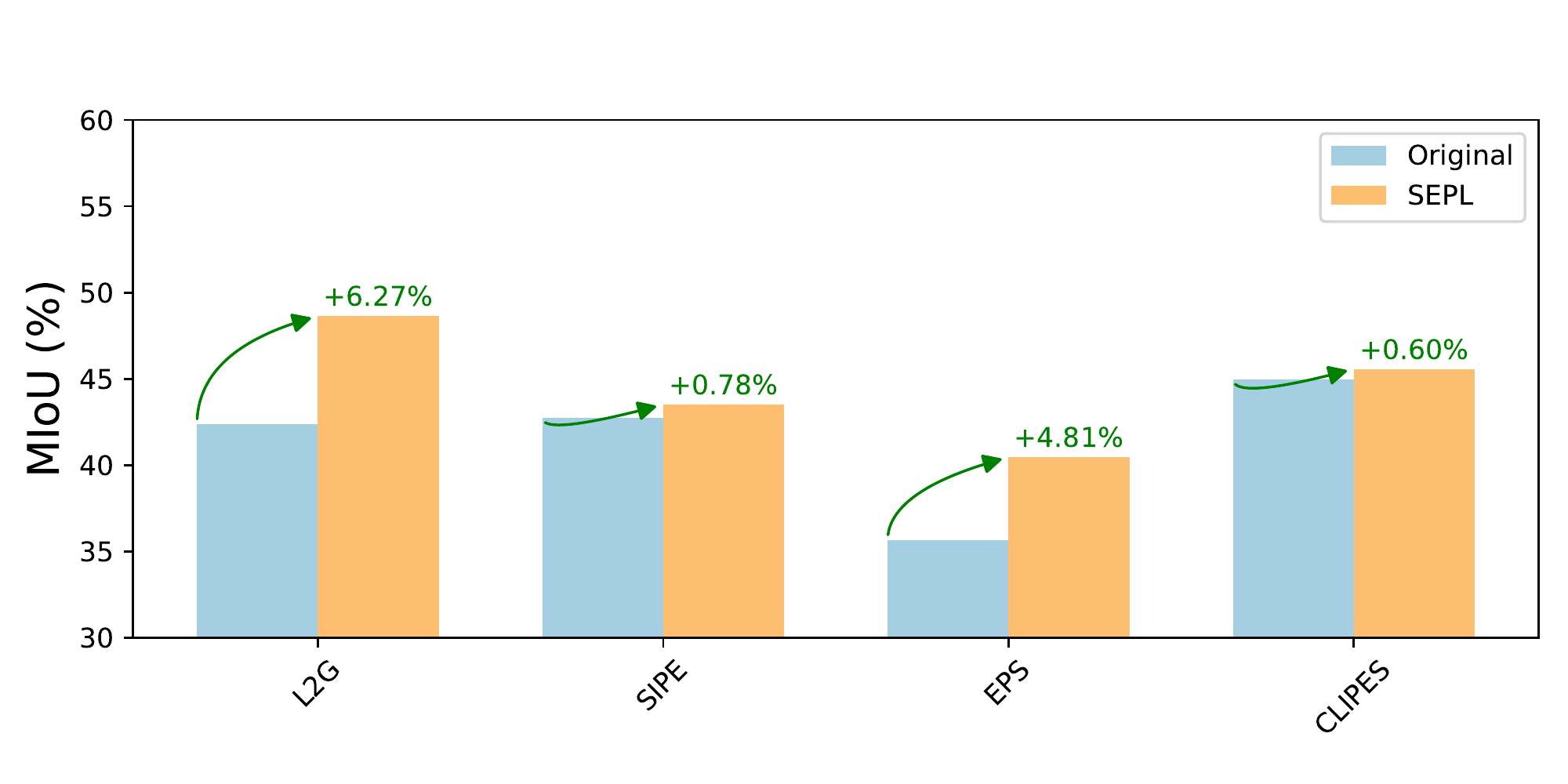}
    \vspace{-8mm}
    \captionof{figure}{\small Pseudo-label quality on COCO $train$ set: Original from WSSS methods vs. SEPL-enhanced. Green indicates enhancements.}
    \label{fig:pseudo_label_coco}
\end{minipage}
\hspace{0.2cm}
\begin{minipage}{.49\textwidth}
\small
    \centering
    \vspace{2.4mm}
    \resizebox{\textwidth}{!}{
    \begin{tabular}{l|ccccc}
    \toprule
    \scriptsize
    Method&{\rotatebox[origin=l]{20}{L2G~\cite{jiang2022l2g}}}& {\rotatebox[origin=l]{20}{SIPE~\cite{SIPE}}} &{\rotatebox[origin=l]{20}{EPS~\cite{lee2021railroad_eps}}}&{\rotatebox[origin=l]{20}{CLIPES~\cite{Lin_2023_CVPR_CLIPES}}}&{\rotatebox[origin=l]{20}{FS~\cite{chen2017deeplabV2}}} \\
    \midrule
    Origin         &43.06 &41.53 &39.06 &46.29 &\multirow{2}{*}{55.04} \\
    SEPL &\textbf{46.39} &\textbf{45.19} &\textbf{41.55} &\textbf{47.90} 
     \\
    \bottomrule
    \end{tabular}}
    \vspace{8.2mm}

    \captionof{table}{\small Reult of Deeplab V2 (ResNet101) without CRF on COCO $val$ set: Original vs. SEPL-enhanced pseudo-labels. FSS: full supervision}
    \label{tab:table2}
\end{minipage}%
\vspace{-5.5mm}

\end{figure}

Figures~\autoref{fig:pseudo_label_voc} and~\autoref{fig:pseudo_label_coco} illustrate the enhanced pseudo-label quality achieved on the PASCAL VOC and MS COCO with our SEPL algorithm. SEPL consistently and significantly elevates the quality of pseudo-labels across various original WSSS methods. Moreover, the impact of this enhancement extends beyond the pseudo-label quality. Utilizing the enhanced pseudo-labels for training supervised semantic segmentation models(DeepLab V2) yields notable improvements in performance. The models, when trained on the enhanced pseudo-labels, consistently outperform those trained on original pseudo-labels. \autoref{tab:table1} and~\autoref{tab:table2} provide a detailed quantitative comparison of these performances across both PASCAL VOC and MS COCO datasets. More detailed results can be found in Appendix~\autoref{app: result}. 
\vspace{-2mm}
\begin{table}[t]
\small
\centering
\resizebox{\textwidth}{!}{
\begin{tabular}{l|cccccccccccccccccccccc}
\toprule
\vspace{-0.5mm}
Method& {\rotatebox[origin=l]{90}{Recurseed~\cite{jo2022recurseed}}}&{\rotatebox[origin=l]{90}{L2G~\cite{jiang2022l2g}}}&{\rotatebox[origin=l]{90}{CLIPES~\cite{Lin_2023_CVPR_CLIPES}}}&{\rotatebox[origin=l]{90}{RCA~\cite{zhou2022regional_rca}}}&{\rotatebox[origin=l]{90}{EPS~\cite{lee2021railroad_eps}}}&{\rotatebox[origin=l]{90}{CLIMS~\cite{xie2022clims}}}&{\rotatebox[origin=l]{90}{TransCAM~\cite{li2023transcam}}}&{\rotatebox[origin=l]{90}{PPC+EPS~\cite{wseg}}}&{\rotatebox[origin=l]{90}{PPC+SEAM~\cite{wseg}}}&{\rotatebox[origin=l]{90}{SIPE~\cite{SIPE}}}&{\rotatebox[origin=l]{90}{PuzzleCAM~\cite{jo2021puzzle}}}&
{\rotatebox[origin=l]{90}{FS~\cite{chen2017deeplabV2}}}\\
\midrule
Origin         &71.38 &69.38 &70.48 &69.48 &68.16 &69.33 &68.10 &70.30 &65.01 &67.14 &65.84 &\multirow{2}{*}{76.48} \\ 
SEPL  &\textbf{72.89} &\textbf{72.41} &\textbf{73.06} &\textbf{69.70} &\textbf{72.13} &\textbf{71.11} &\textbf{69.93} &\textbf{71.93} &\textbf{68.25} &\textbf{69.67} &\textbf{68.89}
 \\
\bottomrule

\end{tabular}}

\vspace{1mm}
\caption{\small Performance of Deeplab V2 (ResNet-101) trained on pseudo-labels without post-processing: Original method vs. SAM-enhanced SEPL. Evaluated on the VOC $val$ set. FS: full supervision}
\vspace{-8mm}
\label{tab:table1}
\end{table}

\vspace{-1.5mm}


\section{When does SAM not help?}
\begin{figure}[t]
    \centering
    \begin{minipage}[b]{0.48\linewidth}
    \scalebox{1}[1.2]
        {\includegraphics[width=\linewidth]{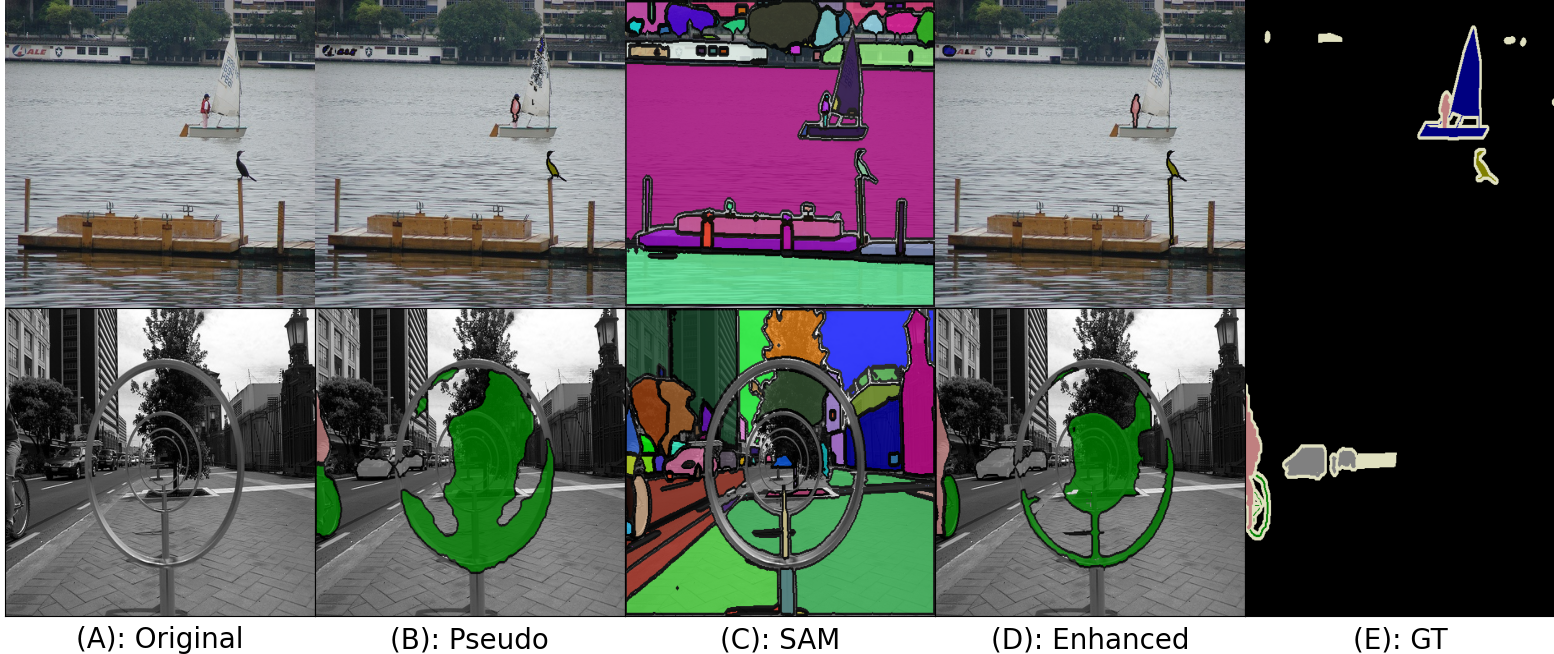}}
        \caption{\small Examples of pseudo-labels activate on incorrect objects or fail to activate on the target objects. Adding SAM mask may detrimentally affect the quality of the pseudo-labels}
        \label{fig:mis_detection_issue}
    \end{minipage}
    \hfill
    \begin{minipage}[b]{0.48\linewidth}
    \scalebox{1}[0.96]{
        \includegraphics[width=\linewidth]{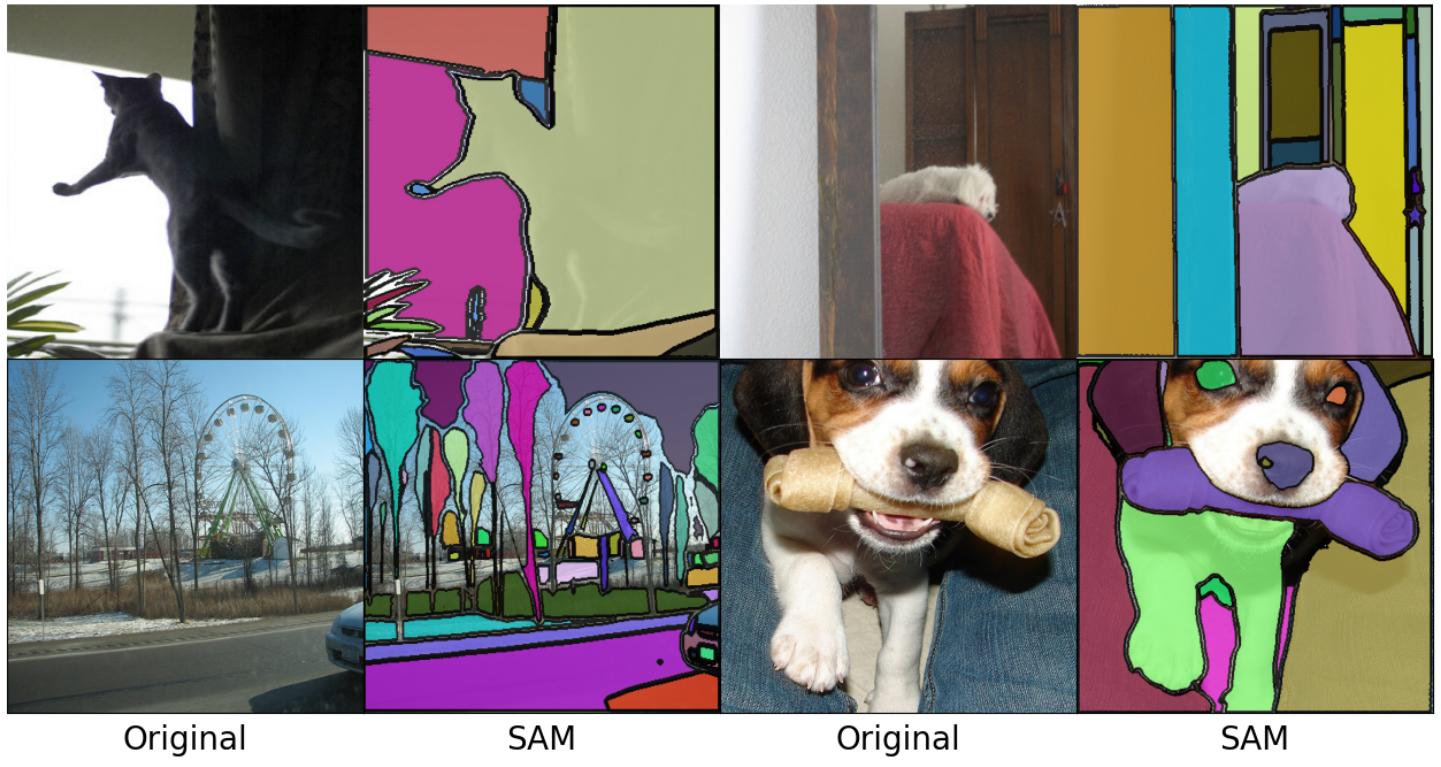}}
        \caption{\small Example of SAM's failures. The first row shows that SAM overlooks certain parts in an image and the second row shows SAM erroneously groups several objects as a single mask}
        \label{fig:sam_fail}
    \end{minipage}
\end{figure}

\begin{figure}[t]
\centering
\vspace{-3.5mm}
\includegraphics[width=1\textwidth]{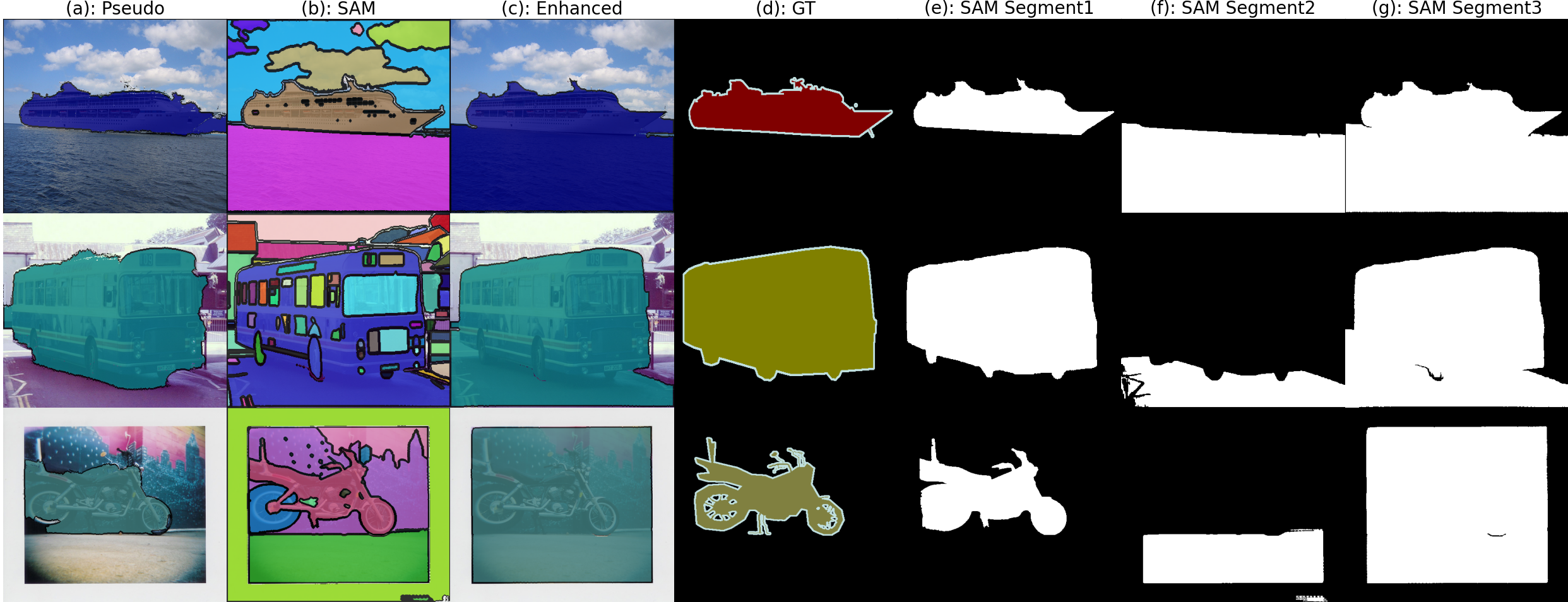}
\caption{\small Certain SAM masks within an image envelop smaller masks. The SAM mask in column (g) entirely covers the masks in columns (e) and (f), which poses a challenge as we cannot ascertain if a mask represents multiple objects or just one. }
\label{fig:non_exclusive_sam}
\vspace{-5mm}
\end{figure}

Upon analyzing instances where SEPL was ineffective, we attributed the shortcomings primarily to three sources: the initial pseudo-labels, the SAM masks, and our enhancement algorithm.
\paragraph{Initial pseudo-labels} Since we leverage the initial pseudo-labels as the anchors to find relevant masks, if they activate on incorrect objects or fail to activate on the target objects, the SAM masks won't offer any enhancement. In fact, they may detrimentally affect the quality of the pseudo-labels, as shown in ~\autoref{fig:mis_detection_issue}. To address this, we turn to better WSSS methods for more precise pseudo-labels.\paragraph{SAM masks} While SAM effectively processes most images in VOC and COCO, it occasionally falters. As depicted in~\autoref{fig:sam_fail}, SAM might sometimes overlook certain parts of images or erroneously group several objects as a single segment. Fine-tuning SAM's inference hyperparameters could be a potential remedy to enhance segmentation outcomes.
\paragraph{Enhancement Algorithm} While our current algorithm is proficient in many scenarios, it falters in specific situations. As depicted in~\autoref{fig:non_exclusive_sam}, certain SAM masks within an image overshadow and envelop smaller masks. An example is column (g) entirely covers the masks in columns (e) and (f). Since pseudo-labels often suffer from partial activation, our existing algorithm inclines towards selecting larger masks to produce a more complete pseudo-label. Yet, this approach poses a dilemma: the absence of clarity on whether these masks encapsulate multiple objects or just a singular entity. As illustrated in the first row of~\autoref{fig:non_exclusive_sam}, the mask in (g) spans both the boat and the sea. Blindly opting for the larger mask risks deteriorating the quality of pseudo-labels. A promising resolution could lie in treating the SAM masks as nodes in a tree, leveraging their inherent hierarchical structure. This tree-based approach might facilitate a more discerning selection of the appropriate masks.

\vspace{-3mm}



\section{Conclusion}
This paper presents a pioneer investigation into the application of SAM as a foundation model in WSSS. By leveraging SAM's class-agnostic capability of producing fine-grained instance masks, we use CAM pseudo-labels as cues to select and combine SAM masks, resulting in high-quality pseudo-labels that are both class-aware and object-aware. Our approach is highly versatile and can be easily integrated into existing WSSS methods without any modification. Despite its simplicity, our approach shows consistent improvement over the SOTA WSSS methods on both PASCAL VOC and MS-COCO datasets. We anticipate that this study will catalyze the adoption of segmentation foundational models across a broad spectrum of computer vision tasks.

\section*{Acknowledgments}
This research is supported in part by NSF (IIS-2107077, OAC2118240, and OAC-2112606) and Cisco Research. We are thankful
for the computational resources of the Ohio Supercomputer Center

{
\small
\printbibliography

@article{asgari2021deep,
  title={Deep semantic segmentation of natural and medical images: a review},
  author={Asgari Taghanaki, Saeid and Abhishek, Kumar and Cohen, Joseph Paul and Cohen-Adad, Julien and Hamarneh, Ghassan},
  journal={Artificial Intelligence Review},
  volume={54},
  pages={137--178},
  year={2021},
  publisher={Springer}
}

@article{lateef2019survey,
  title={Survey on semantic segmentation using deep learning techniques},
  author={Lateef, Fahad and Ruichek, Yassine},
  journal={Neurocomputing},
  volume={338},
  pages={321--348},
  year={2019},
  publisher={Elsevier}
}

@inproceedings{lin2016scribblesup,
  title={Scribblesup: Scribble-supervised convolutional networks for semantic segmentation},
  author={Lin, Di and Dai, Jifeng and Jia, Jiaya and He, Kaiming and Sun, Jian},
  booktitle={Proceedings of the IEEE conference on computer vision and pattern recognition},
  pages={3159--3167},
  year={2016}
}

@article{yuan2021review,
  title={A review of deep learning methods for semantic segmentation of remote sensing imagery},
  author={Yuan, Xiaohui and Shi, Jianfang and Gu, Lichuan},
  journal={Expert Systems with Applications},
  volume={169},
  pages={114417},
  year={2021},
  publisher={Elsevier}
}

@article{feng2020deep,
  title={Deep multi-modal object detection and semantic segmentation for autonomous driving: Datasets, methods, and challenges},
  author={Feng, Di and Haase-Sch{\"u}tz, Christian and Rosenbaum, Lars and Hertlein, Heinz and Glaeser, Claudius and Timm, Fabian and Wiesbeck, Werner and Dietmayer, Klaus},
  journal={IEEE Transactions on Intelligent Transportation Systems},
  volume={22},
  number={3},
  pages={1341--1360},
  year={2020},
  publisher={IEEE}
}

@article{li2023transcam,
  title={Transcam: Transformer attention-based cam refinement for weakly supervised semantic segmentation},
  author={Li, Ruiwen and Mai, Zheda and Zhang, Zhibo and Jang, Jongseong and Sanner, Scott},
  journal={Journal of Visual Communication and Image Representation},
  volume={92},
  pages={103800},
  year={2023},
  publisher={Elsevier}
}

@article{kirillov2023segment,
  title={Segment anything},
  author={Kirillov, Alexander and Mintun, Eric and Ravi, Nikhila and Mao, Hanzi and Rolland, Chloe and Gustafson, Laura and Xiao, Tete and Whitehead, Spencer and Berg, Alexander C and Lo, Wan-Yen and others},
  journal={arXiv preprint arXiv:2304.02643},
  year={2023}
}

@article{jo2022recurseed,
  title={RecurSeed and CertainMix for weakly supervised semantic segmentation},
  author={Jo, Sang Hyun and Yu, In Jae and Kim, Kyung-Su},
  journal={arXiv preprint arXiv:2204.06754},
  year={2022}
}

@article{wang2023seggpt,
  title={Seggpt: Segmenting everything in context},
  author={Wang, Xinlong and Zhang, Xiaosong and Cao, Yue and Wang, Wen and Shen, Chunhua and Huang, Tiejun},
  journal={arXiv preprint arXiv:2304.03284},
  year={2023}
}

@inproceedings{xie2022clims,
  title={CLIMS: cross language image matching for weakly supervised semantic segmentation},
  author={Xie, Jinheng and Hou, Xianxu and Ye, Kai and Shen, Linlin},
  booktitle={Proceedings of the IEEE/CVF Conference on Computer Vision and Pattern Recognition},
  pages={4483--4492},
  year={2022}
}

@inproceedings{wseg,
  title={Weakly supervised semantic segmentation by pixel-to-prototype contrast},
  author={Du, Ye and Fu, Zehua and Liu, Qingjie and Wang, Yunhong},
  booktitle={Proceedings of the IEEE/CVF Conference on Computer Vision and Pattern Recognition},
  pages={4320--4329},
  year={2022}
}

@inproceedings{ru2022learning,
  title={Learning affinity from attention: end-to-end weakly-supervised semantic segmentation with transformers},
  author={Ru, Lixiang and Zhan, Yibing and Yu, Baosheng and Du, Bo},
  booktitle={Proceedings of the IEEE/CVF Conference on Computer Vision and Pattern Recognition},
  pages={16846--16855},
  year={2022}
}

@inproceedings{araslanov2020single,
  title={Single-stage semantic segmentation from image labels},
  author={Araslanov, Nikita and Roth, Stefan},
  booktitle={Proceedings of the IEEE/CVF Conference on Computer Vision and Pattern Recognition},
  pages={4253--4262},
  year={2020}
}

@inproceedings{zhang2020reliability,
  title={Reliability does matter: An end-to-end weakly supervised semantic segmentation approach},
  author={Zhang, Bingfeng and Xiao, Jimin and Wei, Yunchao and Sun, Mingjie and Huang, Kaizhu},
  booktitle={Proceedings of the AAAI Conference on Artificial Intelligence},
  volume={34},
  number={07},
  pages={12765--12772},
  year={2020}
}

@inproceedings{ahn2018learning,
  title={Learning pixel-level semantic affinity with image-level supervision for weakly supervised semantic segmentation},
  author={Ahn, Jiwoon and Kwak, Suha},
  booktitle={Proceedings of the IEEE conference on computer vision and pattern recognition},
  pages={4981--4990},
  year={2018}
}

@inproceedings{ahn2019weakly,
  title={Weakly supervised learning of instance segmentation with inter-pixel relations},
  author={Ahn, Jiwoon and Cho, Sunghyun and Kwak, Suha},
  booktitle={Proceedings of the IEEE/CVF conference on computer vision and pattern recognition},
  pages={2209--2218},
  year={2019}
}

@inproceedings{singh2017hide,
  title={Hide-and-seek: Forcing a network to be meticulous for weakly-supervised object and action localization},
  author={Singh, Krishna Kumar and Lee, Yong Jae},
  booktitle={2017 IEEE international conference on computer vision (ICCV)},
  pages={3544--3553},
  year={2017},
  organization={IEEE}
}

@inproceedings{wei2017object,
  title={Object region mining with adversarial erasing: A simple classification to semantic segmentation approach},
  author={Wei, Yunchao and Feng, Jiashi and Liang, Xiaodan and Cheng, Ming-Ming and Zhao, Yao and Yan, Shuicheng},
  booktitle={Proceedings of the IEEE conference on computer vision and pattern recognition},
  pages={1568--1576},
  year={2017}
}

@inproceedings{li2018tell,
  title={Tell me where to look: Guided attention inference network},
  author={Li, Kunpeng and Wu, Ziyan and Peng, Kuan-Chuan and Ernst, Jan and Fu, Yun},
  booktitle={Proceedings of the IEEE conference on computer vision and pattern recognition},
  pages={9215--9223},
  year={2018}
}

@inproceedings{zhou2016learning,
  title={Learning deep features for discriminative localization},
  author={Zhou, Bolei and Khosla, Aditya and Lapedriza, Agata and Oliva, Aude and Torralba, Antonio},
  booktitle={Proceedings of the IEEE conference on computer vision and pattern recognition},
  pages={2921--2929},
  year={2016}
}

@inproceedings{kolesnikov2016seed,
  title={Seed, expand and constrain: Three principles for weakly-supervised image segmentation},
  author={Kolesnikov, Alexander and Lampert, Christoph H},
  booktitle={Computer Vision--ECCV 2016: 14th European Conference, Amsterdam, The Netherlands, October 11--14, 2016, Proceedings, Part IV 14},
  pages={695--711},
  year={2016},
  organization={Springer}
}

@inproceedings{wang2020self,
  title={Self-supervised equivariant attention mechanism for weakly supervised semantic segmentation},
  author={Wang, Yude and Zhang, Jie and Kan, Meina and Shan, Shiguang and Chen, Xilin},
  booktitle={Proceedings of the IEEE/CVF Conference on Computer Vision and Pattern Recognition},
  pages={12275--12284},
  year={2020}
}

@article{ke2021universal,
  title={Universal weakly supervised segmentation by pixel-to-segment contrastive learning},
  author={Ke, Tsung-Wei and Hwang, Jyh-Jing and Yu, Stella X},
  journal={arXiv preprint arXiv:2105.00957},
  year={2021}
}

@inproceedings{fan2020cian,
  title={Cian: Cross-image affinity net for weakly supervised semantic segmentation},
  author={Fan, Junsong and Zhang, Zhaoxiang and Tan, Tieniu and Song, Chunfeng and Xiao, Jun},
  booktitle={Proceedings of the AAAI Conference on Artificial Intelligence},
  volume={34},
  number={07},
  pages={10762--10769},
  year={2020}
}

@article{choe2020attention,
  title={Attention-based dropout layer for weakly supervised single object localization and semantic segmentation},
  author={Choe, Junsuk and Lee, Seungho and Shim, Hyunjung},
  journal={IEEE transactions on pattern analysis and machine intelligence},
  volume={43},
  number={12},
  pages={4256--4271},
  year={2020},
  publisher={IEEE}
}

@article{dosovitskiy2020image,
  title={An image is worth 16x16 words: Transformers for image recognition at scale},
  author={Dosovitskiy, Alexey and Beyer, Lucas and Kolesnikov, Alexander and Weissenborn, Dirk and Zhai, Xiaohua and Unterthiner, Thomas and Dehghani, Mostafa and Minderer, Matthias and Heigold, Georg and Gelly, Sylvain and others},
  journal={arXiv preprint arXiv:2010.11929},
  year={2020}
}

@article{zou2023segment,
  title={Segment everything everywhere all at once},
  author={Zou, Xueyan and Yang, Jianwei and Zhang, Hao and Li, Feng and Li, Linjie and Gao, Jianfeng and Lee, Yong Jae},
  journal={arXiv preprint arXiv:2304.06718},
  year={2023}
}

@inproceedings{xu2022multi,
  title={Multi-class token transformer for weakly supervised semantic segmentation},
  author={Xu, Lian and Ouyang, Wanli and Bennamoun, Mohammed and Boussaid, Farid and Xu, Dan},
  booktitle={Proceedings of the IEEE/CVF Conference on Computer Vision and Pattern Recognition},
  pages={4310--4319},
  year={2022}
}

@article{everingham2010pascal,
  title={The pascal visual object classes (voc) challenge},
  author={Everingham, Mark and Van Gool, Luc and Williams, Christopher KI and Winn, John and Zisserman, Andrew},
  journal=ijcv,
  volume={88},
  number={2},
  pages={303--338},
  year={2010},
  publisher={Springer}
}

@InProceedings{SIPE,
    author    = {Chen, Qi and Yang, Lingxiao and Lai, Jian-Huang and Xie, Xiaohua},
    title     = {Self-Supervised Image-Specific Prototype Exploration for Weakly Supervised Semantic Segmentation},
    booktitle = {Proceedings of the IEEE/CVF Conference on Computer Vision and Pattern Recognition (CVPR)},
    month     = {June},
    year      = {2022},
    pages     = {4288-4298}
}

@article{SS_survey,
  author={Minaee, Shervin and Boykov, Yuri and Porikli, Fatih and Plaza, Antonio and Kehtarnavaz, Nasser and Terzopoulos, Demetri},
  journal={IEEE Transactions on Pattern Analysis and Machine Intelligence}, 
  title={Image Segmentation Using Deep Learning: A Survey}, 
  year={2022},
  volume={44},
  number={7},
  pages={3523-3542},
  doi={10.1109/TPAMI.2021.3059968}}

@inproceedings{oh2021backgroundBox2,
  title={Background-Aware Pooling and Noise-Aware Loss for Weakly-Supervised Semantic Segmentation},
  author={Oh, Youngmin and Kim, Beomjun and Ham, Bumsub},
  booktitle={CVPR},
  year={2021}
}

@inproceedings{wsss_point1,
  title={Simple does it: Weakly supervised instance and semantic segmentation},
  author={Khoreva, Anna and Benenson, Rodrigo and Hosang, Jan and Hein, Matthias and Schiele, Bernt},
  booktitle=cvpr,
  publisher = {Computer Vision Foundation / {IEEE}},
  pages={876--885},
  year={2017}
}

@inproceedings{wsss_point2,
  title={What’s the point: Semantic segmentation with point supervision},
  author={Bearman, Amy and Russakovsky, Olga and Ferrari, Vittorio and Fei-Fei, Li},
  booktitle=eccv,
  pages={549--565},
  year={2016}
}

@misc{wsss_box1,
      title={BBAM: Bounding Box Attribution Map for Weakly Supervised Semantic and Instance Segmentation}, 
      author={Jungbeom Lee and Jihun Yi and Chaehun Shin and Sungroh Yoon},
      year={2021},
      eprint={2103.08907},
      archivePrefix={arXiv},
      primaryClass={cs.CV}
}

@inproceedings{wsss_box2,
  title={Box-driven class-wise region masking and filling rate guided loss for weakly supervised semantic segmentation},
  author={Song, Chunfeng and Huang, Yan and Ouyang, Wanli and Wang, Liang},
  booktitle=cvpr,
  publisher = {Computer Vision Foundation / {IEEE}},
  pages={3136--3145},
  year={2019}
}

@inproceedings{wsss_box3,
  title={3d guided weakly supervised semantic segmentation},
  author={Sun, Weixuan and Zhang, Jing and Barnes, Nick},
  booktitle={Proceedings of the Asian Conference on Computer Vision},
  year={2020}
}

@inproceedings{wsss1,
  title={Object region mining with adversarial erasing: A simple classification to semantic segmentation approach},
  author={Wei, Yunchao and Feng, Jiashi and Liang, Xiaodan and Cheng, Ming-Ming and Zhao, Yao and Yan, Shuicheng},
  booktitle={Proceedings of the IEEE conference on computer vision and pattern recognition},
  pages={1568--1576},
  year={2017}
}

@inproceedings{wsss2,
  title={Integral object mining via online attention accumulation},
  author={Jiang, Peng-Tao and Hou, Qibin and Cao, Yang and Cheng, Ming-Ming and Wei, Yunchao and Xiong, Hong-Kai},
  booktitle=iccv,
  pages={2070--2079},
  year={2019}
}

@inproceedings{wsss3,
  title={Splitting vs. merging: Mining object regions with discrepancy and intersection loss for weakly supervised semantic segmentation},
  author={Zhang, Tianyi and Lin, Guosheng and Liu, Weide and Cai, Jianfei and Kot, Alex},
  booktitle=eccv,
  year={2020}
  }

@inproceedings{wsss4,
  title={Anti-Adversarially Manipulated Attributions for Weakly and Semi-Supervised Semantic Segmentation},
  author={Lee, Jungbeom and Kim, Eunji and Yoon, Sungroh},
  booktitle=cvpr,
  publisher = {Computer Vision Foundation / {IEEE}},
  pages={4071--4080},
  year={2021}
}

@misc{russakovsky2015imagenet,
      title={ImageNet Large Scale Visual Recognition Challenge}, 
      author={Olga Russakovsky and Jia Deng and Hao Su and Jonathan Krause and Sanjeev Satheesh and Sean Ma and Zhiheng Huang and Andrej Karpathy and Aditya Khosla and Michael Bernstein and Alexander C. Berg and Li Fei-Fei},
      year={2015},
      eprint={1409.0575},
      archivePrefix={arXiv},
      primaryClass={cs.CV}
}

@article{COCO, title={Microsoft Coco: Common Objects in Context}, DOI={10.1007/978-3-319-10602-1_48}, journal={Computer Vision – ECCV 2014}, author={Lin, Tsung-Yi and Maire, Michael and Belongie, Serge and Hays, James and Perona, Pietro and Ramanan, Deva and Dollár, Piotr and Zitnick, C. Lawrence}, year={2014}, pages={740–755}}

@misc{resnet,
      title={Deep Residual Learning for Image Recognition}, 
      author={Kaiming He and Xiangyu Zhang and Shaoqing Ren and Jian Sun},
      year={2015},
      eprint={1512.03385},
      archivePrefix={arXiv},
      primaryClass={cs.CV}
}

@misc{vgg,
      title={Very Deep Convolutional Networks for Large-Scale Image Recognition}, 
      author={Karen Simonyan and Andrew Zisserman},
      year={2015},
      eprint={1409.1556},
      archivePrefix={arXiv},
      primaryClass={cs.CV}
}

@inproceedings{cam,
    author    = {Zhou, Bolei and Khosla, Aditya and Lapedriza, Agata and Oliva, Aude and Torralba, Antonio},
    title     = {Learning Deep Features for Discriminative Localization},
    booktitle = {Computer Vision and Pattern Recognition},
    year      = {2016}
}

@InProceedings{affinitynet,
author = {Ahn, Jiwoon and Kwak, Suha},
title = {Learning Pixel-Level Semantic Affinity With Image-Level Supervision for Weakly Supervised Semantic Segmentation},
booktitle = {Proceedings of the IEEE Conference on Computer Vision and Pattern Recognition (CVPR)},
month = {June},
year = {2018}
}

@misc{saliency1,
      title={Discovering Class-Specific Pixels for Weakly-Supervised Semantic Segmentation}, 
      author={Arslan Chaudhry and Puneet K. Dokania and Philip H. S. Torr},
      year={2017},
      eprint={1707.05821},
      archivePrefix={arXiv},
      primaryClass={cs.CV}
}

@article{yao2020saliency2,
  title={Saliency guided self-attention network for weakly and semi-supervised semantic segmentation},
  author={Yao, Qi and Gong, Xiaojin},
  journal={IEEE Access},
  year={2020}
}

@inproceedings{salency2,
  title={Railroad is not a Train: Saliency as Pseudo-pixel Supervision for Weakly Supervised Semantic Segmentation},
  author={Lee, Seungho and Lee, Minhyun and Lee, Jongwuk and Shim, Hyunjung},
  booktitle={CVPR},
  year={2021}
}

@inproceedings{chen2014semantic_Deeplabv1,
  author    = {Liang{-}Chieh Chen and
               George Papandreou and
               Iasonas Kokkinos and
               Kevin Murphy and
               Alan L. Yuille},
  title     = {Semantic Image Segmentation with Deep Convolutional Nets and Fully Connected CRFs},
  booktitle = {ICLR},
  year      = {2015}
}

@article{chen2017deeplabV2,
  title={Deeplab: Semantic image segmentation with deep convolutional nets, atrous convolution, and fully connected crfs},
  author={Chen, Liang-Chieh and Papandreou, George and Kokkinos, Iasonas and Murphy, Kevin and Yuille, Alan L},
  journal={TPAMI},
  year={2017}
}

@inproceedings{jo2021puzzle,
  title={Puzzle-cam: Improved localization via matching partial and full features},
  author={Jo, Sanghyun and Yu, In-Jae},
  booktitle={2021 IEEE International Conference on Image Processing (ICIP)},
  pages={639--643},
  year={2021},
  organization={IEEE}
}

@inproceedings{lee2021railroad_eps,
  title={Railroad is not a train: Saliency as pseudo-pixel supervision for weakly supervised semantic segmentation},
  author={Lee, Seungho and Lee, Minhyun and Lee, Jongwuk and Shim, Hyunjung},
  booktitle={Proceedings of the IEEE/CVF conference on computer vision and pattern recognition},
  pages={5495--5505},
  year={2021}
}

@inproceedings{jiang2022l2g,
  title={L2g: A simple local-to-global knowledge transfer framework for weakly supervised semantic segmentation},
  author={Jiang, Peng-Tao and Yang, Yuqi and Hou, Qibin and Wei, Yunchao},
  booktitle={Proceedings of the IEEE/CVF conference on computer vision and pattern recognition},
  pages={16886--16896},
  year={2022}
}

@InProceedings{Lin_2023_CVPR_CLIPES,
    author    = {Lin, Yuqi and Chen, Minghao and Wang, Wenxiao and Wu, Boxi and Li, Ke and Lin, Binbin and Liu, Haifeng and He, Xiaofei},
    title     = {CLIP Is Also an Efficient Segmenter: A Text-Driven Approach for Weakly Supervised Semantic Segmentation},
    booktitle = {Proceedings of the IEEE/CVF Conference on Computer Vision and Pattern Recognition (CVPR)},
    month     = {June},
    year      = {2023},
    pages     = {15305-15314}
}

@inproceedings{zhou2022regional_rca,
  title={Regional semantic contrast and aggregation for weakly supervised semantic segmentation},
  author={Zhou, Tianfei and Zhang, Meijie and Zhao, Fang and Li, Jianwu},
  booktitle={Proceedings of the IEEE/CVF Conference on Computer Vision and Pattern Recognition},
  pages={4299--4309},
  year={2022}
}

@inproceedings{lin2014microsoft,
  title={Microsoft coco: Common objects in context},
  author={Lin, Tsung-Yi and Maire, Michael and Belongie, Serge and Hays, James and Perona, Pietro and Ramanan, Deva and Doll{\'a}r, Piotr and Zitnick, C Lawrence},
  booktitle={Computer Vision--ECCV 2014: 13th European Conference, Zurich, Switzerland, September 6-12, 2014, Proceedings, Part V 13},
  pages={740--755},
  year={2014},
  organization={Springer}
}

@article{chen2017deeplab,
  title={Deeplab: Semantic image segmentation with deep convolutional nets, atrous convolution, and fully connected crfs},
  author={Chen, Liang-Chieh and Papandreou, George and Kokkinos, Iasonas and Murphy, Kevin and Yuille, Alan L},
  journal={IEEE transactions on pattern analysis and machine intelligence},
  volume={40},
  number={4},
  pages={834--848},
  year={2017},
  publisher={IEEE}
}
}
\clearpage
\appendix
\section*{\LARGE Appendix}
\section{Related Work}
\label{ssup_related}


\subsection{Weakly Supervised Semantic Segmentation (WSSS)}

Recent approaches in weakly supervised semantic segmentation (WSSS) often rely on Class Activation Maps (CAM)~\cite{zhou2016learning} to  generate pixel-level pseudo-labels. These pseudo-labels are then used to train the segmentation model in a fully supervised manner. However, CAM often exhibits a bias towards the most discriminative regions of the target object which limits the quality of the pseudo-labels. To overcome this challenge, recent works mainly focus on generating high-quality CAMs with integral activation on the entire object regions. 

Early-stage works~\cite{singh2017hide,wei2017object,li2018tell} encourage the network to discover less activated object parts via adversarial erasing. In addition to the classification loss typically used in the WSSS framework, specific loss functions such as SEC loss~\cite{kolesnikov2016seed}, equivariance regularization~\cite{wang2020self}, and contrastive loss~\cite{ke2021universal,wseg} have been exploited in narrowing the gap between the pixel-level and image-level supervisions. Some works also introduce network modules to address the partial activation problem of CAM: SEAM~\cite{wang2020self} leverages pixel-level semantic affinities with a pixel correlation module; CIAN~\cite{fan2020cian} exploits the additional information from related images with a cross-image affinity module. Recent methods based on Vision Transformer~\cite{dosovitskiy2020image} including~\cite{li2023transcam, xu2022multi} aim to uncover more comprehensive object regions by exploring the global information from the attention of the transformer network. Most of these works follow the multi-stage framework, where a post-processing step is necessary for refining and improving the initial pseudo-labels generated from CAM.

\subsection{Post-Processing in WSSS}
\label{sec: post}
Although there are end-to-end WSSS solutions~\cite{ru2022learning,araslanov2020single,zhang2020reliability} available, most of the recent works still rely on some post-processing techniques to enhance the initial pseudo-labels to achieve superior performance. Among these techniques, two widely utilized methods for refining pseudo-labels are AffinityNet~\cite{ahn2018learning} and IRNet~\cite{ahn2019weakly}. AffinityNet trains a network from CAM to predict the semantic affinities and uses it for propagating local activations, whereas IRNet~\cite{ahn2019weakly} learns and predicts semantic affinities more effectively by leveraging class boundary maps. Despite the substantial improvement in the pseudo-labels, these methods require training a separate network. The computational cost involved can be a significant barrier, especially when working with large-scale datasets. Additionally, the careful tuning of hyperparameters to obtain accurate foreground and background pixels for training these networks can slow down the entire training pipeline. This requirement for meticulous parameter tuning not only adds complexity to the process but also limits the applicability and scalability of these post-processing techniques.

\section{Experiment Details}

\subsection{Dataset Details}
\label{app: data}
Experiments are conducted on two publicly available datasets, PASCAL VOC 2012~\cite{everingham2010pascal}and MS COCO 2014~\cite{COCO}. The PASCAL VOC 2012 dataset contains 20 semantic categories and the background. It is split into three sets, the training, validation, and test sets, each containing 1464, 1449, and 1456 images, respectively. Following the standard setting, we also use the augmented training set, yielding a total of 10582 training images. The MS COCO 2014 dataset has 80 semantic categories. Following~\cite{choe2020attention}, the images without target categories are excluded from the dataset, remaining 82081 training images and 40137 validation images. We report the mean Intersection-over-Union (mIoU), precision, and recall for evaluation. To demonstrate the quality116
of the pseudo-labels, we evaluate them on the VOC and COCO training set.

\subsection{Baselines}
\label{app: baseline}
\begin{itemize}
    \item \textbf{CLIMS:}~\cite{xie2022clims} A Cross-Language Image Matching framework leveraging natural language supervision to activate complete object regions and suppress related open background regions for improved CAM quality in WSSS.
    \item \textbf{SIPE:}~\cite{SIPE} Self-supervised Image-specific Prototqualitativeype Exploration, which tailors prototypes for each image to capture complete regions, optimizing feature representation and enabling self-correction for improved WSSS performance.
    \item \textbf{PPC:}~\cite{wseg} A weakly-supervised pixel-to-prototype contrast method providing pixel-level supervisory signals, executed across and within different views of an image to enhance the quality of pseudo masks for WSSS.
    \item \textbf{TransCAM:}~\cite{li2023transcam} A Conformer-based solution that refines CAM by leveraging attention weights from the transformer branch of the Conformer, capturing both local features and global representations for WSSS.
    \item \textbf{RecurSeed:}~\cite{jo2022recurseed} An approach that alternately reduces non-detections and false-detections through recursive iterations, implicitly finding an optimal junction and leveraging a novel data augmentation method, EdgePredictMix, for improved WSSS performance.
    \item \textbf{L2G}~\cite{jiang2022l2g} A simple online local-to-global knowledge transfer framework for high quality object attention mining. It first leverages a local classification network to extract attentions from multiple local patches randomly cropped from the input image. Then,
it utilizes a global network to learn complementary attention knowledge across multiple local attention maps online.
    \item \textbf{CLIPES}~\cite{Lin_2023_CVPR_CLIPES} An approach that leverages CLIP to improve pseudo-label generation, refinement and final segmentation model training.
    \item \textbf{RCA}~\cite{zhou2022regional_rca} RCA is
equipped with a regional memory bank to store massive, diverse object patterns appearing in training data, which acts
as strong support for exploration of dataset-level semantic
structure. 
\item \textbf{EPS}~\cite{lee2021railroad_eps}  EPS learns from pixel-level feedback by combining two weak supervisions; the image-level label provides
the object identity via the localization map and the saliency
map from the off-the-shelf saliency detection model offers
rich boundaries
\item \textbf{PuzzleCAM}~\cite{jo2021puzzle} PuzzleCAM minimizes differences between the features from separate patches and the whole image. It consists of a puzzle module and two regularization terms to discover the most integrated region in an object. 
\end{itemize}
\subsection{Implementation Details}
\label{app: impl}
\paragraph{SAM Inference Hyperparameters} For our experiments, we adopted the standard settings of the SAM model as provided in their official repository. However, we made modifications to two specific hyperparameters to tailor the model's behavior to our needs:

\begin{itemize}
\item --pred-iou-thresh was set from None to 
0.86.
\item --stability-score-thresh was set from None to 
0.92.
\end{itemize}

By adjusting these thresholds, our objective was to enable SAM to produce a wider and more diverse range of masks for selection via our algorithm. Importantly, these modifications did not have a detrimental effect on the inference speed, ensuring efficiency was maintained throughout the process.
For inference, we employed the default pretrained ViT-H SAM model.

\section{Extra Results}
\label{app: result}

\begin{figure}[h]
\centering
\begin{subfigure}[b]{0.3\linewidth}
\includegraphics[width=\linewidth]{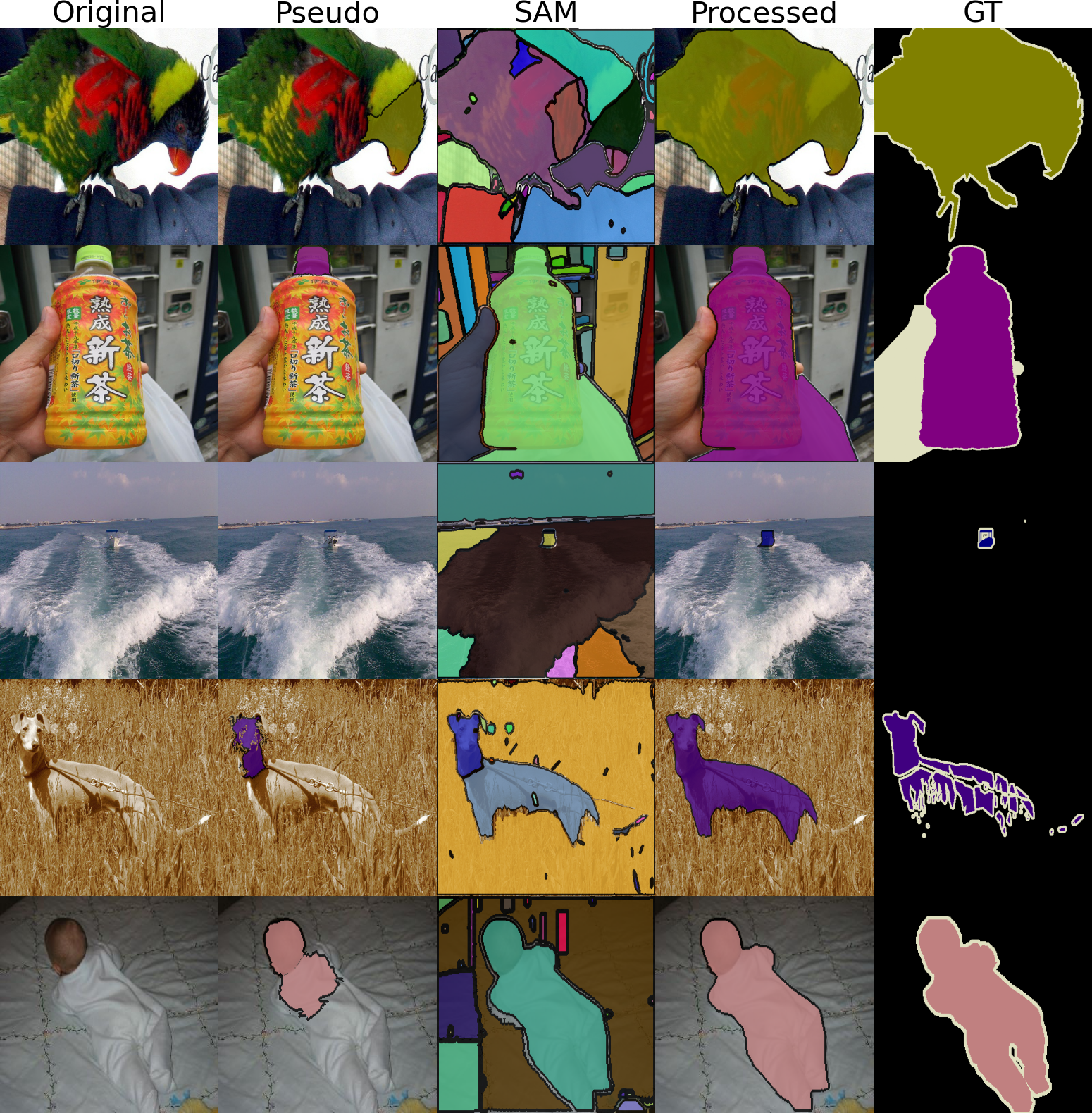}
\caption{Recall}
\label{fig:CAM_postprocessing_recall}
\end{subfigure}
\begin{subfigure}[b]{0.3\linewidth}
\includegraphics[width=\linewidth]{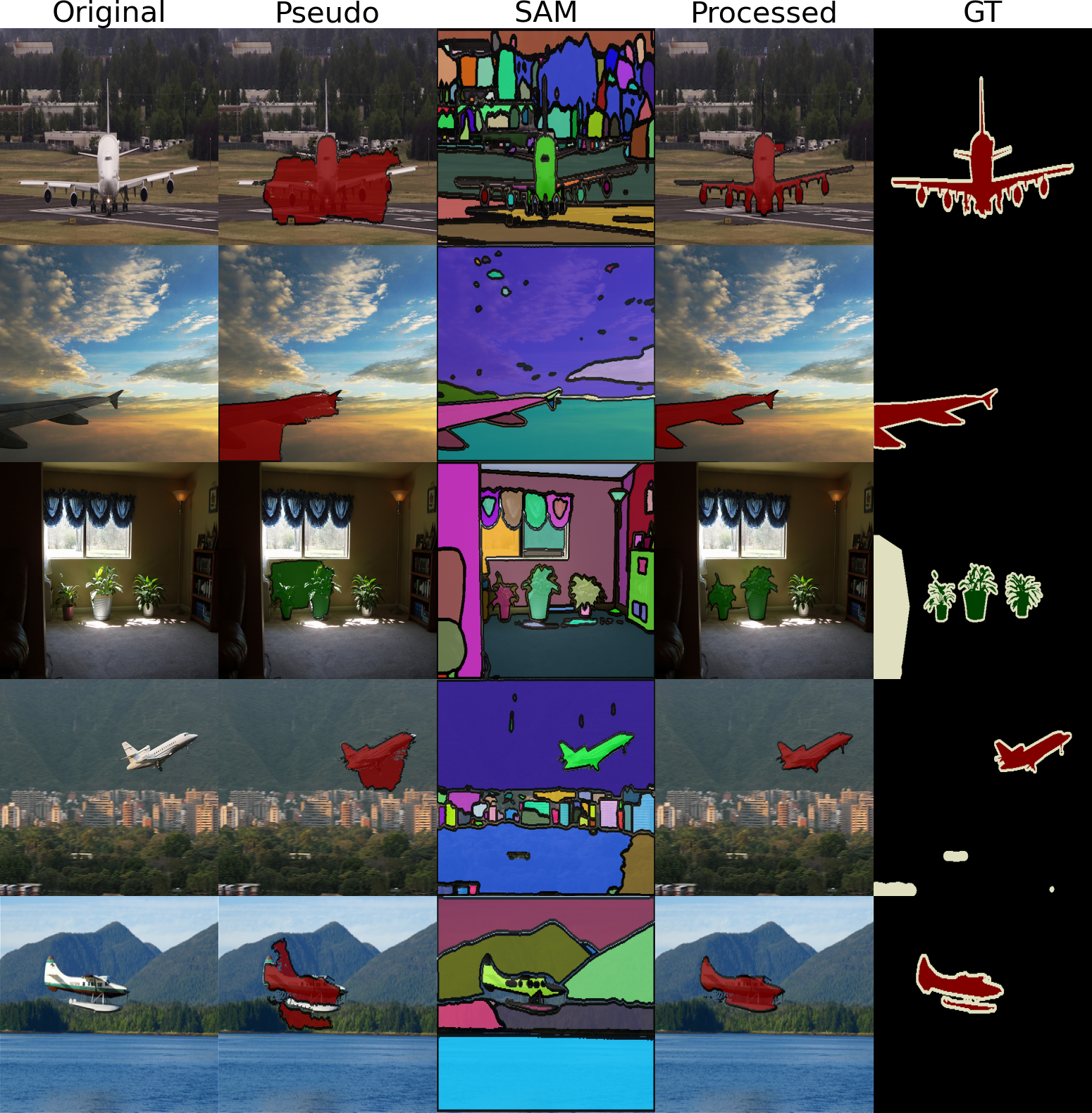}
\caption{Precision}
\label{fig:CAM_postprocessing_precision}
\end{subfigure}
\begin{subfigure}[b]{0.3\linewidth}
\includegraphics[width=\linewidth]{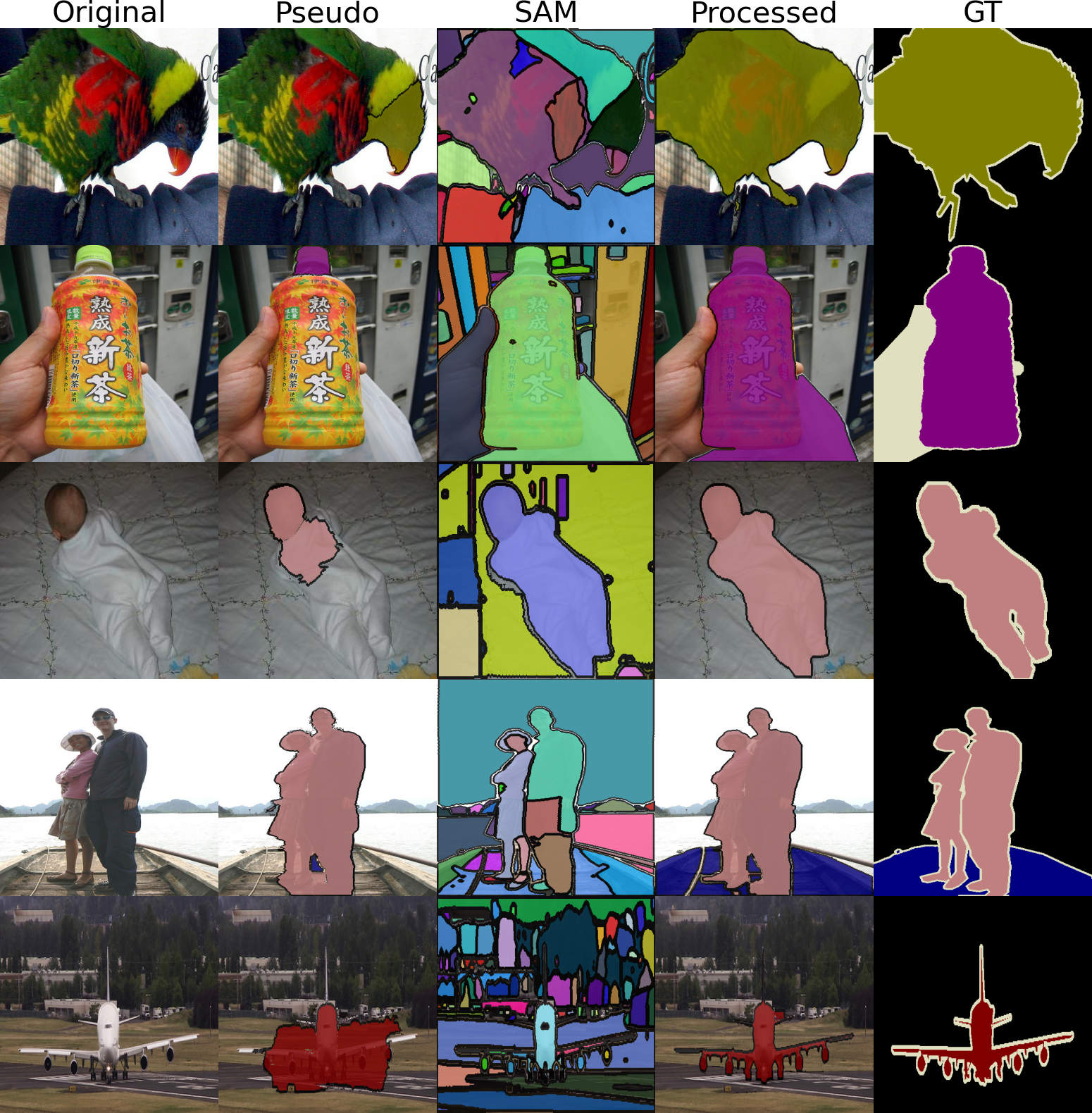}
\caption{mIoU}
\label{fig:CAM_postprocessing_mIoU}
\end{subfigure}
\caption{Improvements for pseudo-labels: (a) Recall, (b) Precision, (c) mIoU.}
\label{fig:CAM_postprocessing}
\end{figure}

\autoref{fig:CAM_postprocessing} illustrates the qualitative improvements of the pseudo-labels enhanced by SEPL in recall, precision, and mIoU, respectively, which are calculated based on the average of all samples. \autoref{tab:pre_rec} shows the quantitative improvement of the pseudo-labels enhanced by SEPL for recall and precision. 

\begin{table}[h]
\large
\centering
\resizebox{\textwidth}{!}{
\begin{tabular}{l|ccccccccccccccccccccc}
\toprule
Method& {\rotatebox[origin=l]{90}{Recurseed}}&{\rotatebox[origin=l]{90}{L2G}}&{\rotatebox[origin=l]{90}{CLIPES}}&{\rotatebox[origin=l]{90}{RCA}}&{\rotatebox[origin=l]{90}{EPS}}&{\rotatebox[origin=l]{90}{CLIMS}}&{\rotatebox[origin=l]{90}{TransCAM}}&{\rotatebox[origin=l]{90}{PPC+EPS}}&{\rotatebox[origin=l]{90}{PPC+SEAM}}&{\rotatebox[origin=l]{90}{SIPE}}&{\rotatebox[origin=l]{90}{PuzzleCAM}} \\
\midrule
Pseudo Label Precision         &85.10 &78.86 &83.81 &81.53 &78.10 &80.97 &80.46 &82.73 &78.93& 76.75&83.29 \\
SEPL Precision &\textbf{85.73} &\textbf{84.57} &\textbf{85.96} &\textbf{82.13} &\textbf{84.85} &\textbf{83.29} &\textbf{81.89} &\textbf{84.32} &\textbf{81.76} &\textbf{79.86} &\textbf{83.93}
 \\
 Precision Delta &\color{red}{0.63} &\color{red}{5.71} &\color{red}{2.15} &\color{red}{0.60} &\color{red}{6.75} &\color{red}{2.32} &\color{red}{1.43} &\color{red}{1.59} &\color{red}{2.83} &\color{red}{3.11} &\color{red}{6.39}\\
 \hline
 Pseudo Label Recall         &85.85 &87.35 &85.36 &71.04 &84.93 &83.19 &84.83&86.91 &83.45 &87.68 &76.77\\
SEPL Recall &\textbf{91.75} &\textbf{92.05} &\textbf{91.58} &\textbf{79.37} &\textbf{90.94} &\textbf{89.12} &\textbf{90.07} &\textbf{92.02} &\textbf{90.36} &\textbf{91.12} &\textbf{86.22}\\
  Recall Delta&\color{red}{5.90} &\color{red}{4.70} &\color{red}{6.22} &\color{red}{8.33} &\color{red}{6.01} &\color{red}{5.93} &\color{red}{5.24} &\color{red}{5.11} &\color{red}{6.91} &\color{red}{3.44} &\color{red}{9.45}\\
\bottomrule

\end{tabular}}
\vspace{1mm}
\caption{Pseudo label precision and recall improvements by incorporating SEPL}
\label{tab:pre_rec}
\end{table}

\subsection{Apply SAM on CAM without post-processing}
As mentioned in Section~\autoref{sec: post}, most recent works still rely on some post-processing techniques to enhance the initial CAM, aiming to procure more precise pseudo-labels. However, these enhancement procedures often demand substantial computational overhead and extended training durations. Such constraints can potentially hinder the broad-scale deployment of WSSS on extensive datasets. Since the initial CAM also provides an estimation of object localization, our proposed SEPL can be directly applied to the initial CAM. This approach circumvents the need for additional post-processing steps, leading to appreciable reductions in both training duration and computational demands in the WSSS pipeline.

\begin{figure}[tb]
\centering
    \includegraphics[width=0.95\linewidth]{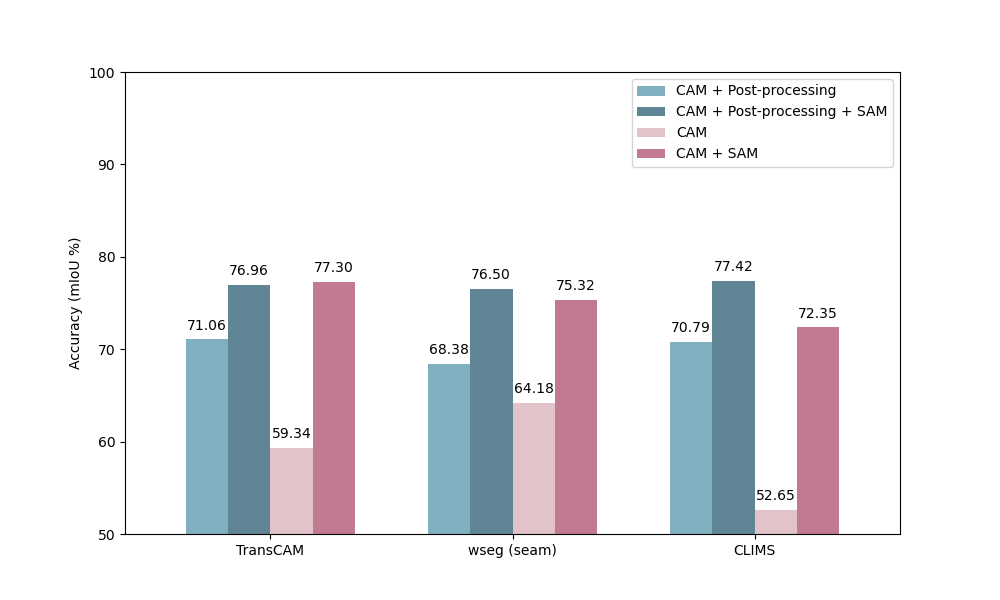}
    \caption{ By directly utilizing iinitial CAM with SAM,
we achieved comparable performance to that of post-processed pseudo-labels enhanced by SAM.
This finding suggests that SAM can be used as a substitute for post-processing modules, resulting in
a marked acceleration of the entire WSSS training pipeline}
    \label{fig:pp_res}
\end{figure}

\begin{figure}[H]
 \vspace{10mm}
\centering
\begin{subfigure}[b]{0.3\linewidth}
\includegraphics[width=\linewidth]{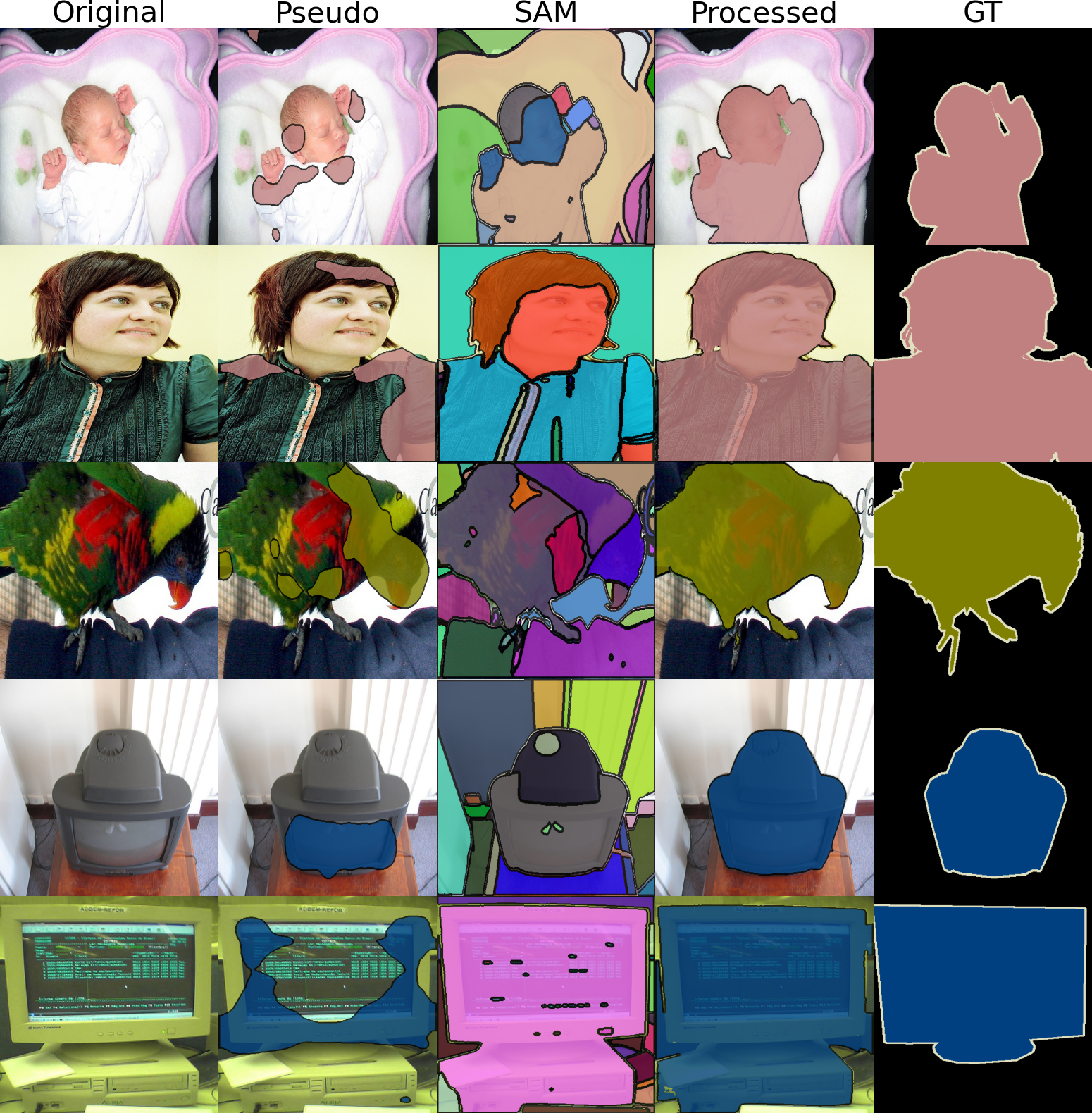}
\caption{Recall}
\label{fig:CAM_without_postprocessing_recall}
\end{subfigure}
\begin{subfigure}[b]{0.3\linewidth}
\includegraphics[width=\linewidth]{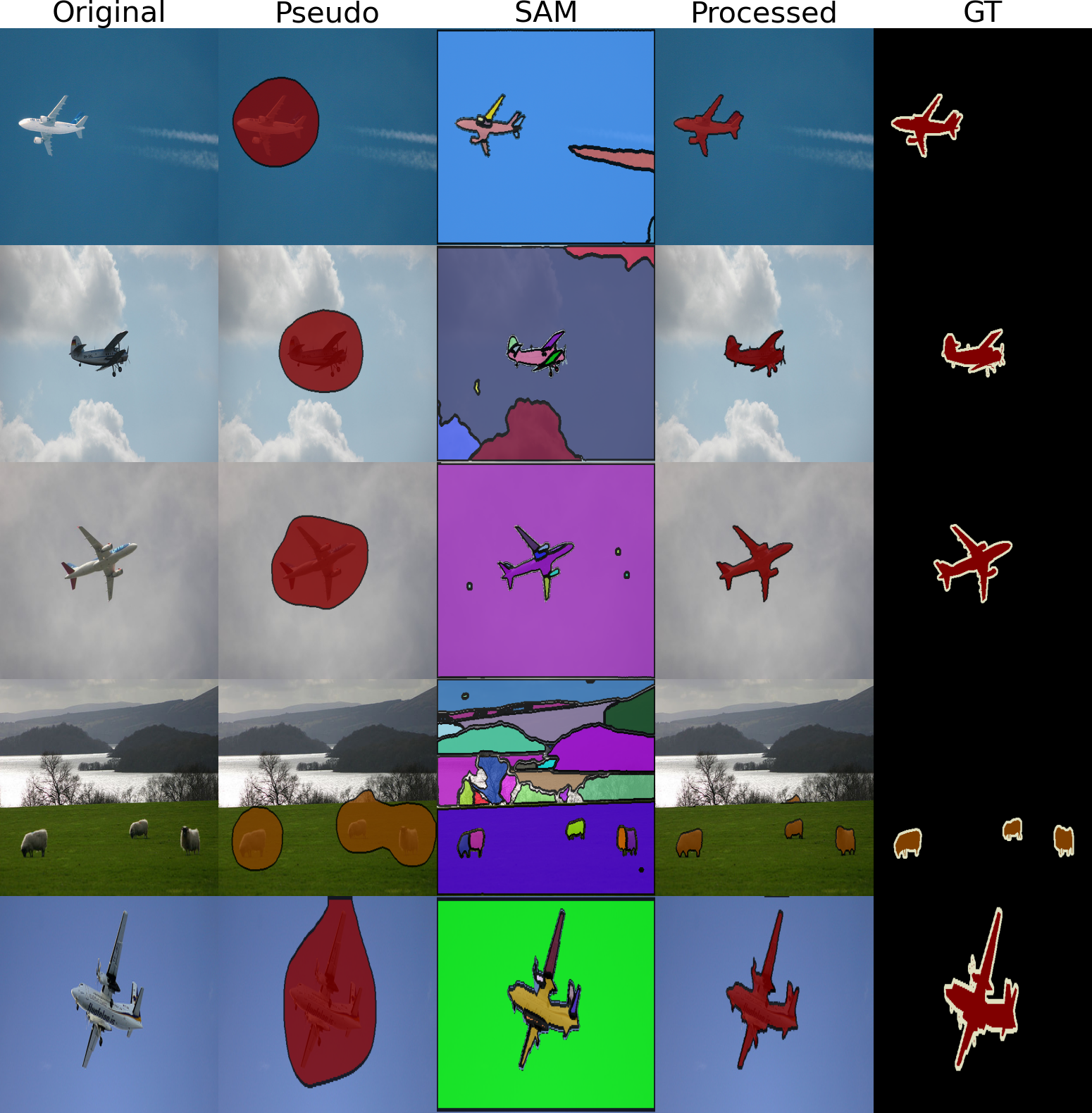}
\caption{Precision}
\label{fig:CAM_without_postprocessing_precision}
\end{subfigure}
\begin{subfigure}[b]{0.3\linewidth}
\includegraphics[width=\linewidth]{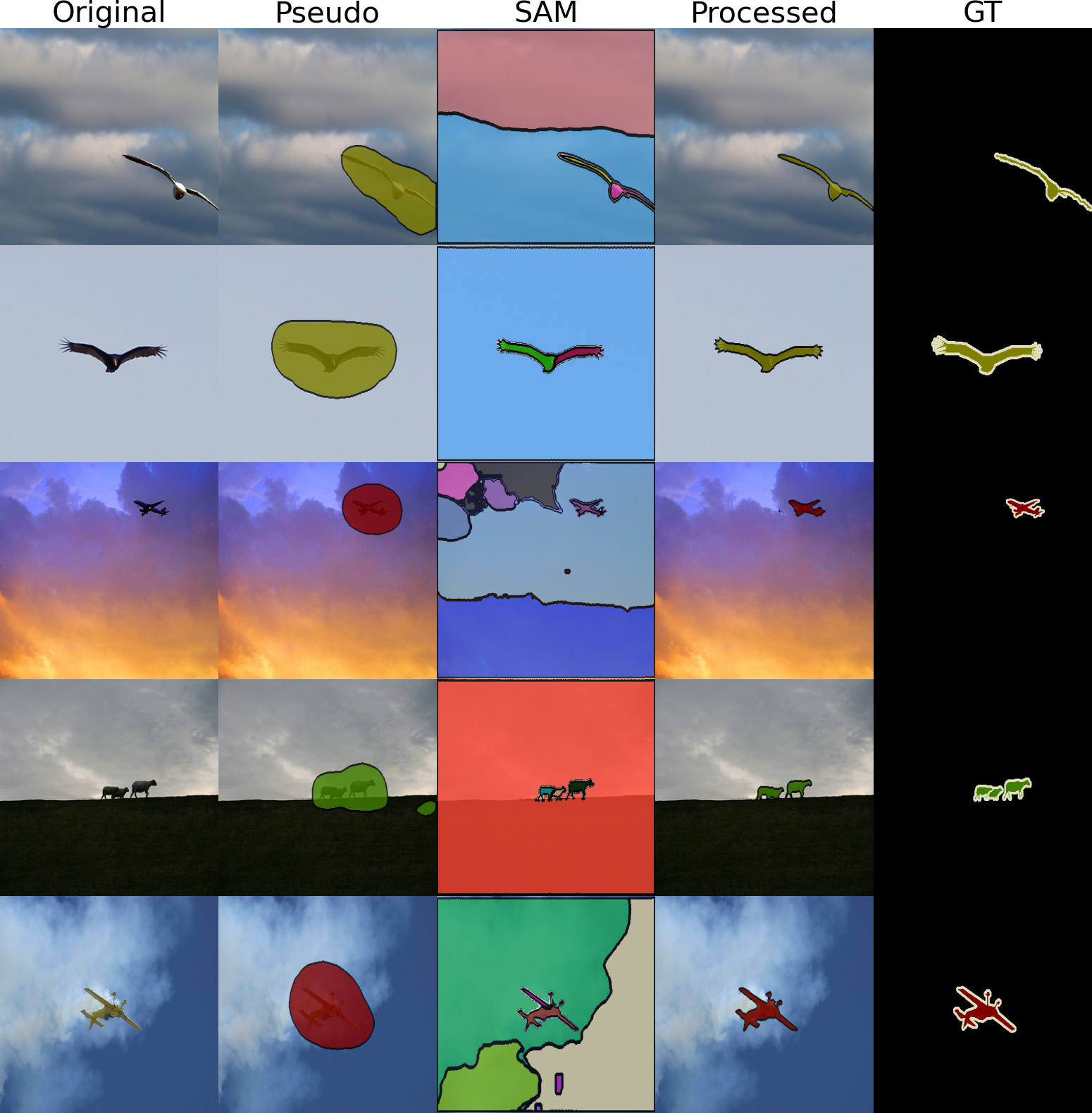}
\caption{mIoU}
\label{fig:CAM_without_postprocessing_mIoU}
\end{subfigure}
\caption{Improvements for CAM without post-processing: (a) Recall, (b) Precision, (c) mIoU.}
\label{fig:CAM_without_postprocessing}
\end{figure}

As shown in~\autoref{fig:pp_res}, the initial CAM consistently benefits from SAM masks. More interestingly, CAM+SAM can reach and even surpass the quality of pseudo-labels obtained after conventional post-processing. This finding suggests that SAM has the potential to replace time-consuming post-processing steps, offering a more efficient solution to WSSS tasks. \autoref{fig:CAM_without_postprocessing} illustrates the qualitative improvements of the initial CAM by SEPL in recall, precision, and mIoU, respectively. 

\end{document}